\documentclass[final]{clv2025}

\jvol{vv}
\jnum{nn}
\jyear{2025}
\dochead{Short Paper} 
\pageonefooter{Action editor: Deyi Xiong. Submission received: 6 July 2024; revised version received: 25 May 2025; accepted for publication: 7 July 2025.}
\usepackage{amsmath}
\usepackage{arydshln}
\usepackage{color}
\usepackage{xcolor}
\usepackage{hyperref}
\usepackage{multirow}
\usepackage{multicol}
\usepackage{booktabs}
\usepackage{amssymb}
\usepackage{lingmacros}
\usepackage{soul}
\usepackage{ragged2e}
\usepackage{bbding}
\usepackage{appendix}
\usepackage{xspace}
\definecolor{darkblue}{rgb}{0, 0, 0.5}
\newcommand{\meta}{Meta4XNLI\xspace}
\newcommand{\metaes}{Meta4XNLI\textsubscript{\langid{ES}}\xspace}
\newcommand{\metaen}{Meta4XNLI\textsubscript{\langid{EN}}\xspace}
\newcommand{\esxnli}{esXNLI\xspace}
\newcommand{\xnli}{XNLI\xspace}
\newcommand{\xnlitest}{XNLI\textsubscript{test}\xspace}
\newcommand{\xnlidev}{XNLI\textsubscript{dev}\xspace}
\newcommand{\cometa}{CoMeta\xspace}
\newcommand{\vuam}{VUAM\xspace}
\newcommand{\mdeberta}{mDeBERTa\xspace}
\newcommand{\xlmroberta}{XLM-RoBERTa\xspace}
\newcommand{\llamasm}{Llama-3.1-8B-Instruct\xspace}
\newcommand{\qwensm}{Qwen2.5-7B-Instruct\xspace}
\newcommand{\gemma}{gemma-7b-it\xspace}
\newcommand{\llamalg}{Llama-3.3-70B-Instruct\xspace}
\newcommand{\qwenlg}{Qwen2.5-72B-Instruct\xspace}
\newcommand{\gpt}{gpt-4o\xspace}

\newcommand{\langid}[1]{\texttt{#1}\xspace}
\runningtitle{Metaphor in XNLI}

\runningauthor{Sanchez-Bayona and Agerri}

\begin{document}

\title{Meta4XNLI: A Cross-lingual Parallel Corpus for Metaphor Detection and Interpretation}

%\author{Elisa Sanchez-Bayona elisa.sanchez@ehu.eus}
%\affil{HiTZ Basque Center for Language Technology - Ixa NLP Group, University of the Basque Country UPV/EHU}

%\author{Rodrigo Agerri rodrigo.agerri@ehu.eus}
%\affil{HiTZ Basque Center for Language Technology - Ixa NLP Group, University of the Basque Country UPV/EHU}

\author{Elisa Sanchez-Bayona\thanks{Corresponding author}$^{1}$, Rodrigo Agerri$^{1}$}

\affilblock{
\affil{HiTZ Basque Center for Language Technology - Ixa, University of the Basque Country UPV/EHU\\\quad \email{\{elisa.sanchez, rodrigo.agerri\}@ehu.eus}}
}

\maketitle

\begin{abstract}

Metaphors are a ubiquitous but often overlooked part of everyday language. As a complex cognitive-linguistic phenomenon, they provide a valuable means to evaluate whether language models can capture deeper aspects of meaning, including semantic, pragmatic, and cultural context. In this work, we present \meta, the first parallel dataset for Natural Language Inference (NLI) newly annotated for metaphor detection and interpretation in both English and Spanish. Meta4XNLI facilitates the comparison of encoder- and decoder-based models in detecting and understanding metaphorical language in multilingual and cross-lingual settings.
Our results show that fine-tuned encoders outperform decoders-only LLMs in metaphor detection. Metaphor interpretation is evaluated via the NLI framework with comparable performance of masked and autoregressive models, which notably decreases when the inference is affected by metaphorical language. Our study also finds that translation plays an important role in the preservation or loss of metaphors across languages, introducing shifts that might impact metaphor occurrence and model performance. These findings underscore the importance of resources like \meta for advancing the analysis of the capabilities of language models and improving our understanding of metaphor processing across languages. Furthermore, the dataset offers previously unavailable opportunities to investigate metaphor interpretation, cross-lingual metaphor transferability, and the impact of translation on the development of multilingual annotated resources.

\end{abstract}

\section{Introduction}

Metaphor is commonly characterized as the understanding of an abstract concept in terms of another concept from a more concrete domain. According to \cite{Lakoff80metaphorswe}, we can establish a distinction between \textbf{conceptual metaphors}, cognitive mappings that arise from the association between source and target domains, and \textbf{linguistic metaphors}, the expression of these mappings through language. The pervasiveness of metaphors in our daily speech makes it fundamental for language models to be able to process them accordingly, in order to achieve a satisfactory interaction between users and these tools. In addition, metaphor processing may have implications for other Natural Language Processing (NLP) tasks such as Machine Translation \cite{mao-etal-2018-word, Schffner2004MetaphorAT, shutova2013statistical}, political discourse analysis \cite{Charteris-Black2011, prabhakaran2021metaphors, rodriguez2023paper} or hate speech \cite{lemmens2021improving}, among others. Since we study metaphor occurrence in natural language sentences, in this work, we will focus on linguistic metaphors only. 

The task most explored to date is \textbf{metaphor detection} or identification, typically framed as a sequence labeling problem grounded on different theoretical proposals \cite{wilks1975preferential, wilks1978making, searle1979expression, black1962models}. The most popular methodology is perhaps the one defined by the MIPVU guidelines \cite{steenetal2010}, which relies on the mismatch between the basic and contextual meaning of a potential metaphor. The application of this procedure resulted in the publication of the referential dataset \vuam. As usual, most published work is English-centered, although multilingual and cross-lingual approaches are increasingly gaining popularity. However, resources for other languages are still scarce, of reduced size, or automatically labeled and non-parallel. 

\begin{figure}[h!]
\resizebox{\textwidth}{!}{
\begin{tabular}{lll}
\toprule
\textbf{Example} & \textbf{NLI tag} & \textbf{Met} \\ \midrule
\textbf{Premise1}: \textit{You are very outgoing and \textbf{open} with the fans.} &  &  \\ \midrule
\textbf{H1}: \textit{You meet with your fans after each concert.} & Neu & Yes \\
\textbf{H2}: \textit{You have a really good relationship with the fans.} & Ent & Yes \\ 
\textbf{H3}: \textit{You ignore your fans.} & Contra & Yes \\ \midrule \midrule
\textbf{Premise2}: \textit{And, she didn't really understand.} &  &  \\ \midrule 
\textbf{H1}: \textit{Alas, she was not able to understand \textbf{clearly} due to a language \textbf{barrier}.} & Neu & Yes\\
\textbf{H2}: \textit{Indeed, she did not comprehend.} & Ent & No \\ 
\textbf{H3}: \textit{She knew exactly what we were talking about.} & Contra & No\\ 
\bottomrule
\end{tabular}}
\caption{\label{tab:meta_examples}
Examples from \meta with annotations. Tokens (in \textbf{bold}) in premises and hypotheses are labeled for metaphor detection. Column \textit{Met} represents annotations for interpretation. For premises or hypotheses containing metaphorical expressions, we marked those pairs in which the understanding of the metaphor is essential to infer the right relation (``ent'': entailment, ``neu'': neutral, ``contra'': contradiction).}
\end{figure}

Although \textbf{metaphor interpretation} has been less explored than detection, there is a growing interest in recent years, reflected, for example, in the celebration of the FigLang 2022 Shared Task on Understanding Figurative Language \cite{saakyan-etal-2022-report}. Previous work commonly approached understanding as a paraphrasing task \cite{shutova2010automatic, Shutova2012UnsupervisedMP, shutova2013statistical, Shutova2013MetaphorIA, bizzoni-lappin-2018-predicting}. However, most recent works frame it within the task of Natural Language Inference (NLI), which consists of determining the relationship between a premise and a hypothesis, generally entailment, neutral, or contradiction \cite{agerri2008metaphor, chakrabarty-etal-2021-figurative, stowe-etal-2022-impli, chakrabarty-etal-2022-flute, kabra-etal-2023-multi}. 

Nevertheless, most of the datasets published in this particular paradigm are limited to one language, typically English, or consist of premises and hypotheses that are identical except for the metaphorical expressions which are replaced by their literal or antonym counterparts to construct entailment and contradiction pairs \cite{agerri2008metaphor, mohler2013applying, chakrabarty-etal-2021-figurative, stowe-etal-2022-impli, chakrabarty-etal-2022-flute, kabra-etal-2023-multi}. Although this lexical substitution mechanism to develop datasets might appear fruitful, these artifacts do not represent uses of metaphorical expressions in natural language occurring text.

This work addresses these shortcomings by exploring metaphor detection as a sequence labeling task and metaphor interpretation via NLI from a multilingual and cross-lingual perspective by leveraging a parallel dataset with newly added annotations for both tasks. Our contributions can be summarized as follows:

\begin{itemize}
    \item We provide \meta (M4X), a collection of existing NLI datasets, \xnli \cite{conneau-etal-2018-xnli} and \esxnli \cite{artetxe-etal-2020-translation}, enriched with metaphor annotations for the tasks of detection and interpretation in Spanish (\langid{ES}) and English (\langid{EN}). To the best of our knowledge, this is the first multilingual parallel dataset with metaphor annotations for both tasks in naturally occurring text. \meta's main features include:
    \begin{itemize}
        \item 13K parallel sentences with metaphorical annotations at the token and premise-hypothesis pair levels. An example of our dataset is shown in Figure \ref{tab:meta_examples}.
        \item Supports cross-lingual analysis of metaphorical language use and transfer.
        \item Metaphor annotations across texts of multiple domains and natural language sentences.
        \item Enables the study of how metaphorical expressions influence a natural language understanding task, such as NLI, providing a foundation for evaluating the metaphor interpretation abilities of language models.
        \item Contains texts translated in both directions: \langid{EN} -> \langid{ES}, \langid{ES} -> \langid{EN}, facilitating analysis of how translation affects metaphor preservation and interpretation.
    \end{itemize}
    
    \item Monolingual and cross-lingual experiments in various evaluation setups by leveraging \meta: 
    \begin{itemize}
    \item For metaphor detection, we trained models in other datasets and evaluated them cross-domain in \meta; we fine-tuned and evaluated Masked Language Models (MLM, encoders only) and decoders-only Large Language Models (LLMs) with \meta for both languages and also in zero-shot scenarios. Through the usage of different datasets and a cross-lingual approach, we aim to explore the generalization capabilities of Language Models (LM) and the extent of knowledge transfer when it comes to metaphor processing. Our results show a more competitive performance of MLMs, while LLMs struggle with the task.
    \item Concerning metaphor interpretation, we framed the task within NLI to study the capabilities of Language Models (LMs to refer to both MLMs and LLMs) to understand metaphorical language. First, we tested MLMs and LLMs fine-tuned for the task of NLI with metaphorical and non-metaphorical pairs from \meta. Second, we also trained and evaluated MLMs on the whole dataset. In contrast to previous work \cite{rakshit2023does,stowe-etal-2022-impli,chakrabarty-etal-2022-flute}, our results demonstrate that LMs obtain worse results in NLI whenever metaphorical expressions affect the inferential task.
    \end{itemize}
    \item Code and data publicly available\footnote{\url{https://huggingface.co/datasets/HiTZ/meta4xnli}}.
\end{itemize}

\section{Related Work} \label{sec:related_work}

In this section, we present an overview of the most significant works focused on metaphor processing. First, we analyze metaphor detection and interpretation developed in \langid{EN}, since most previous works are English-centered. Afterwards, we discuss multi- and cross-lingual approaches for both tasks.

\paragraph{\textbf{Metaphor Detection}}

Initially, the majority of the work on metaphor detection was corpus-based \cite{charteris2004corpus, semino2017corpus}. However, the growing interest over the last years led to the celebration of FigLang shared tasks \cite{reportleong-etal-2018-report, leong-etal-2020-report}, which promoted multiple approaches to address it as a sequence labeling task addressed with deep learning techniques. The most popular methodology consisted of training a model with specific features, either linguistically related \cite{stowe-palmer-2018-leveraging} or of other nature, such as abstractness, visual, or emotion-related information \cite{tsvetkov2014metaphor, bizzoni2018bigrams, tong-etal-2021-recent, neidlein-etal-2020-analysis}.

The arrival of Transformer-based models \cite{devlin2018bert} led to huge improvements in this task. Most fine-tuned models are founded on linguistic theories, such as MIP \citep{steenetal2010} or \emph{Selectional Preference} (SP) \cite{wilks1975preferential, percy1958metaphor}, which, generally speaking, address metaphor as a contrast between basic and contextual meaning. The MIP approach was extended to MIPVU to develop the reference corpus in \langid{EN}: the VUAM dataset \cite{steenetal2010}, which covers texts of multiple domains and annotations at token level, and was used for the shared tasks along with the release of TOEFL dataset \cite{leong-etal-2020-report}. Other well-known datasets are TroFi \cite{birke2006clustering}, considerably smaller in size and restricted to verbs; or the MOH-X dataset \cite{mohammad-etal-2016-metaphor}, which also focuses on metaphorical and literal examples of verbs. Other available corpora cover texts from a single domain, such as tweets \cite{Zayed2019CrowdSourcingAH} or news headlines, in NewsMet dataset \cite{joseph-etal-2023-newsmet}.

The combination of pre-trained language models and these available resources brought forth multiple models fine-tuned for metaphor detection with \textit{ad hoc} architectures. For instance, \citet{song-etal-2021-verb} propose Mr-BERT, a model capable of extracting the grammatical and semantic relations of a metaphorical verb and its context. RoPPT \cite{wang-etal-2023-metaphor-roppt}  takes into account information from dependency trees to extract the terms most relevant to the target word. The purpose of other published models is to identify the metaphoric span of the sentence, namely MelBERT \cite{melbert}, based on MIP and SP theories, as well as BasicBERT \cite{li-etal-2023-metaphor-meanings-model}. To alleviate the scarcity of metaphor-annotated data, CATE \cite{lin-etal-2021-cate} is a ContrAstive Pre-Trained ModEl that uses semi-supervised learning and self-training.

Others use additional linguistic resources besides datasets, like \citet{robertawilde}, which takes advantage of Wiktionary definitions to build their MLM MIss RoBERTa WiLDe; FrameBERT \cite{li-etal-2023-framebert} uses FrameNet \cite{fillmore-etal-2002-framenet} to extract the concept of the detected metaphor. MisNET \cite{zhang-liu-2022-metaphor} exploits dictionary resources and is based on linguistic theories to predict word-level metaphors. The model of \citet{wan-etal-2021-enhancing} learns from glosses of the definition of the contextual meaning of metaphors. \citet{maudslay-teufel-2022-metaphorical} present what they call a \emph{Metaphorical Polysemy Detection} model by exploiting WordNet and Word Sense Disambiguation (WSD) to perform the detection. Another approach is to frame metaphor detection within another NLP task, such as \citet{zhang-liu-2023-adversarial}, who adopt a multi-task learning framework where knowledge from WSD is leveraged to identify metaphors; \citet{feng-ma-2022-better} apply an Auto-Augmented Structure-aware generative model that approaches metaphor detection as a keywords-extraction task; and the work of \citet{dankin-etal-2022-yes}, which explores few-shot scenarios from a Yes-No Question-Answering perspective. \citet{badathala-etal-2023-match} also propose a multi-task approach to detect metaphor and hyperbole, although at the sentence level. 

Among these examples, the state of the art on the task evaluated as sequence labeling on VUA-20 was the following: DeBERTa-large (73.79 F1) fine-tuned and evaluated on VUAM \cite{sanchez-bayona-agerri-2022-leveraging}, BasicBERT (73.3 F1), FrameBERT (73.0), RoPPT (72.8 F1) and MelBERT (72.3 F1). These scores show the complexity of the task and how there is still room for improvement to achieve competitive performance.
 
\paragraph{\textbf{Metaphor Interpretation}}
The evaluation of metaphor interpretation remains a difficult, open research problem, which is why most works frame it within other NLU tasks, namely, paraphrasing or NLI. Among the works based on paraphrasing, we can find both supervised \cite{shutova2013statistical, Shutova2013MetaphorIA} and unsupervised \cite{shutova2010automatic, Shutova2012UnsupervisedMP} approaches. The work of \citet{bollegala2013metaphor} explores the generation of literal paraphrases for metaphorical verbs in an unsupervised manner. \citet{bizzoni-lappin-2018-predicting} developed a corpus with sentences that contain metaphorical expressions and a set of literal paraphrases ranked according to their acceptability. They exploit their resource to test deep learning systems that approach metaphor interpretation as a classification and ranking task. \citet{mao2021interpreting} also focused on paraphrasing of verbal metaphors as well. They take advantage of MOH-X \cite{mohammad-etal-2016-metaphor} and \vuam to test BERT's capability to generate the most probable literal substitute. \citet{pedinotti-etal-2021-howling} provide an evaluation dataset in \langid{EN} with 300 instances that include conventional and novel metaphors, as well as literal and nonsense sentences. They exploit it to test BERT's ability to interpret metaphors and discriminate among the different types of sentences, in addition to examining how MLMs encode this knowledge. 

Some recent works address metaphor interpretation as a Question-Answering problem. They reformulate metaphorical expressions as questions or prompts to test LLMs. \citet{comsa-etal-2022-miqa} propose MiQA, a dataset of 300 items that gather literal and metaphorical premises, paired with implication sentences to evaluate LLMs' metaphor understanding by asking if the implications are true or false. \citet{liu-etal-2022-testing} develop Fig-QA dataset also for \langid{EN} but of a considerably larger size. It comprises 10,256 instances of creative metaphors paired with their literal implication sentences. They exploit their resource to evaluate state-of-the-art models' ability to understand metaphor, framed as an inference task. Nevertheless, these pairs are not natural utterances, but human-generated examples that present a fixed sentence structure. In addition, premises and their implications present high lexical overlapping, which could create a bias in the models' evaluation results, e.g., \emph{Money is a helpful stranger} - > \emph{Money is good}; \emph{Money is a murderer} - > \emph{Money is bad}. 

\citet{rakshit2023does} introduce FigurativeQA, which gathers 1000 yes/no questions in \langid{EN} from the reviews domain that include metaphorical and literal examples in addition to other figurative language phenomena, such as simile, hyperbole, idiom, and sarcasm to probe models. Other works like \citet{wachowiak-gromann-2023-gpt, pitarch-etal-2023-mean} center their research on examining whether generative models understand conceptual metaphors and their reasoning capabilities.

A popular approach that continues to be explored is the study of metaphor interpretation framed in the NLI or Recognizing Textual Entailment (RTE) task \cite{agerri2008metaphor, mohler2013applying}. For example, \citet{chakrabarty-etal-2021-figurative} propose 12500 instances in \langid{EN} collected from existing datasets for RTE, which cover simile, metaphor, and irony. Of the total dataset, only 600 pairs are metaphor-related. In addition, their data is generated by lexical substitution, which does not truthfully represent the use of metaphors in natural language. For instance, entailment pairs of hypotheses are generated by the literal substitution of the metaphorical expression from the premise. This also occurs with non-entailment. In this case, the metaphor from the premise is replaced by an antonym to generate the hypothesis.

The work of \citet{zayed-etal-2020-figure} aimed to create a gold standard for metaphor interpretation. They developed a dataset of 2500 tweets with definitions of verb-noun metaphorical expressions with the aid of lexical resources and word embeddings. However, they do not present any experiments on models to evaluate their dataset.
\citet{stowe-etal-2022-impli} introduce IMPLI, a semi-automatic constructed dataset for \langid{EN} to evaluate RoBERTa-like models' performance on figurative language, specifically idioms and metaphors, using NLI as an evaluation framework. Their resource is considerably larger (24,029 silver and 1,831 gold sentence pairs) and covers entailment and non-entailment (a merge of neutral and contradiction) relations.

As is the case with other previously reviewed datasets, some biases are introduced during the generation of this dataset. Entailed pairs only consist of sentences that include a metaphorical expression, while the entailed hypothesis corresponds to its literal paraphrase. Furthermore, non-entailment pairs were developed by ``creating'' a literal but not natural meaning of the figurative expression. 
\citet{chakrabarty-etal-2022-flute} propose FLUTE, an explanation-based dataset in \langid{EN} of 9000 NLI pairs that include sarcasm, simile, metaphor, and idioms. FLUTE differs from the other resources in that each entailed/contradicted pair is accompanied by an explanation. However, premises and hypotheses are based on lexical substitution: e.g., \emph{His dark clothes had a negative effect in the shadows} -> \emph{His dark clothes were a plus in the shadows}.

The limited availability of metaphor interpretation datasets in languages other than English, combined with the templatic structure of existing resources, presents a significant obstacle to a generalizable evaluation of the capabilities of language models to understand metaphorical language. Many current datasets rely on lexical substitution methods, which can introduce artifacts that might favor model performance \cite{artetxe-etal-2020-translation, naik-etal-2018-stress}. To advance metaphor understanding in multilingual contexts and balance the prevalence of English resources, it is necessary to develop parallel datasets with metaphor annotations in naturally occurring text across more than one language.

\paragraph{\textbf{Multilingual Approaches}}

% Multilingual work
The related work discussed so far is centered on English only. In the case of metaphor detection, and to compensate for the lack of language diversity, some monolingual datasets have been published in other languages. However, their size is rather limited and they tend to be monolingual, e.g., the KOMET corpus \cite{antloga2020komet} in Slovene, \cometa \cite{sanchez-bayona-agerri-2022-leveraging} in Spanish, Estonian \cite{aedmaa-etal-2018-combining}, German \cite{koper-schulte-im-walde-2016-distinguishing} or Polish adjective-noun metaphors \cite{mykowiecka-etal-2018-literal}. 

Among the datasets that cover more than one language, we can find LCC \cite{mohammad-etal-2016-metaphor}: it comprises texts of English, Spanish, Russian, and Farsi and provides source/target, metaphoricity degree, and metaphor/literal annotations on sentences. However, there are no labels at the token level, annotations were extracted automatically, only a subset was manually validated, and it is not parallel text. The work of \citet{levin-etal-2014-resources} proposes the CCM corpora, which also includes sentences in the same four languages as the LCC dataset. Nevertheless, it is composed of source and target mappings. Thus, it is centered on conceptual metaphors. 
\citet{schuster-markert-2023-nut} explores static embeddings to generate a cross-lingual dataset out of existing resources in German, English, and Polish, but focuses on adjective-noun metaphor pairs and non-parallel texts. Similarly, \citet{berger-2022-transfer} explores transfer learning techniques, such as neural machine translation, cross-lingual word embeddings, or multilingual pre-trained language models, to obtain a dataset for metaphor detection in German from English corpora. The work of \citet{wang-etal-2024-mmte} presents a parallel dataset in Chinese and English. However, it is not annotated for any of the tasks of our interest.

In addition to data generation, the experimental settings of many recent publications have shifted to multi- and/or cross-lingual approaches, with the objective of analysing metaphor transferability among different languages. Previous to the appearance of pre-trained language models, \citet{tsvetkov2014metaphor} already explored unsupervised methods to evaluate metaphor identification of specific syntactic constructions in English, Spanish, Russian, and Farsi. \citet{shutova2017multilingual} also experimented with semi- and unsupervised learning, specifically with clustering techniques for metaphor detection in English, Spanish, and Russian.
\citet{probing-aghazadeh-etal-2022-metaphors} investigate whether pre-trained language models are able to encode metaphorical meaning through the task of metaphor detection in English, Farsi, Russian, and Spanish.

The work of \citet{multi-multifigurative} presents a combination of joint models able to detect whether a sentence contains hyperbole, idioms, or metaphorical expressions in these same four languages. Nevertheless, it does not provide annotations for the text at span level. 

Concerning metaphor interpretation, the range of available datasets in multiple languages is more limited than for detection.
Most of the data already mentioned is available only in English: FigurativeQA \cite{rakshit2023does}, Fig-QA \cite{liu-etal-2022-testing}, IMPLI \cite{stowe-etal-2022-impli}, or FLUTE \cite{chakrabarty-etal-2022-flute}. The exception is the work of \citet{kabra-etal-2023-multi}, who address figurative language understanding from a multi-lingual and multi-cultural perspective. They present MABL, a dataset for seven languages with a high number of speakers, such as Hindi, Swahili, or Sudanese. They demonstrate that socio-cultural features essentially impact conceptual mappings that later materialize in linguistic metaphors. They evaluate language models on MABL and provide some insights regarding the English- or Western-centered training process of these models. 

To summarize, this overview gives an account of the prevalence of \langid{EN} in metaphor processing research. While some monolingual detection datasets exist for other languages, they are quite small when compared to those in \langid{EN}. In fact, any fairly sized corpora for metaphor detection different from English do not include annotations at the token level or are limited to metaphorical expressions of a given POS. An example is the LCC dataset, which is a popular non-parallel dataset that only includes annotations at the sentence level.

Datasets annotated for metaphor interpretation are even more scarce, and the lack of variety of languages is more pronounced than in metaphor detection corpora. The only multilingual dataset is MABL. However, neither the data nor the annotations are parallel across languages. Furthermore, while it is possible to find different datasets, these corpora tend to include biases or artifacts \cite{boisson2023construction}. Thus, in paraphrasing datasets, tuples are commonly composed of a sentence with a metaphorical expression that is replaced with its literal meaning. In the case of NLI datasets, premise-hypothesis pairs are typically constructed by lexical substitution: the entailment is based on the literal paraphrase of the metaphor, whereas the contradiction is obtained by replacing the metaphorical expression with its antonym. Therefore, these instances are not representative of spontaneous occurrences of metaphor in natural language text. 

% Please add the following required packages to your document preamble:
\begin{table}[h!]
\caption{Available datasets for metaphor detection and interpretation according to their features; \textbf{task}: \textit{det} for detection, \textit{paraphr} for paraphrase; \textbf{Level}: Annotation level, namely, sentence or token level; \textbf{NL}: spontaneous language, in opposition to sentences with a fixed structure or generated through lexical substitution; \textbf{$*$}: corpus structured in sentence pairs, usually hypothesis and premise in NLI.}
\label{tab:dataset_comparison}
\footnotesize
\begin{tabular}{p{2.5cm}p{1.9cm}lcccccr}
\toprule
                        \textbf{Dataset}    &  \textbf{Lang}  & \textbf{Task} & \textbf{Level} &  \textbf{POS}   & \textbf{Parallel} & \textbf{\#sents} & \textbf{NL}  \\ \toprule
\vuam    & \langid{EN}  & det   & tok &  all    & \XSolidBrush        & 16202  & \Checkmark    \\ 
\citet{berger-2022-transfer}  & \langid{DE}, \langid{EN}  & det   & tok     &  all     & \Checkmark        &   500      & \Checkmark   \\
\cometa   & \langid{ES}  & det   & tok &  all      & \XSolidBrush   & 3633  & \Checkmark   \\
MOH-X  & \langid{EN}  & det   & tok &  verb      & \XSolidBrush     & 647  & \Checkmark   \\
\citet{mohler-etal-2016-introducing}  & \langid{EN}  & det   & tok &  all    & \XSolidBrush   & 1450  & \Checkmark   \\
\citet{badathala-etal-2023-match}  & \langid{EN}  & det   & sent &  -      & \XSolidBrush        & 16024 & \Checkmark   \\
TroFi & \langid{EN}  & det   & tok &  verb      & \XSolidBrush        & 3713  & \Checkmark   \\
LCC     & \langid{EN}, \langid{ES}, \langid{FA}, \langid{RU}  & det   & sent &  -   & \XSolidBrush    & 40138 & \Checkmark \\ 
TOEFL    & \langid{EN}  & det   & tok &  all      & \XSolidBrush    & 3709  & \Checkmark   \\  
KOMET & \langid{SL} & det   & tok &  all       & \XSolidBrush    & 13963  & \Checkmark   \\ 
Aedmaa et al. (2018) & \langid{ET} & det   & tok &  verb       & \XSolidBrush    & 2000  & \Checkmark   \\ 
Mykowiecka et al. (2018) & \langid{PL} & det   & sent &  adj, noun   & \XSolidBrush    & 2052  & \Checkmark   \\ 
Zayed et al (2019) & \langid{EN} & det   & sent &  verb, noun   & \XSolidBrush    & 2500  & \Checkmark   \\
Köper et al. (2016) & \langid{DE} & det   & token &  adj, noun   & \XSolidBrush    & 6436  & \Checkmark   \\ \midrule
\multicolumn{8}{c}{\textit{Interpretation}} \\ \midrule
\citet{bizzoni-lappin-2018-predicting} & \langid{EN} & paraphr   & sent &  -       & \XSolidBrush    & 1000 & \Checkmark   \\ 
FLUTE & \langid{EN} & NLI   & sent &  -      & \XSolidBrush    & $^*$1500 & \XSolidBrush   \\ 
Fig-QA  & \langid{EN} & QA   & sent &  -     & \XSolidBrush    & 10250  & \XSolidBrush   \\ 
Figurative-QA & \langid{EN} & QA   & sent &  -      & \XSolidBrush    & 400 & \XSolidBrush   \\ 
Figure Me Out & \langid{EN} & paraphr   & tok &  verb, noun      & \XSolidBrush    & 2500 & \Checkmark   \\ 
MiQA  & \langid{EN} & QA   & sent &  -      & \XSolidBrush    & 300  & \Checkmark   \\ 
IMPLI  & \langid{EN} & NLI   & sent &  -    & \XSolidBrush    & $^*$1045  & \XSolidBrush   \\ 
NewsMet  & \langid{EN} & paraphr   & tok, sent &  verb    & \XSolidBrush    & 1519  & \Checkmark   \\ 
MABL  & \langid{HI}, \langid{ID}, \langid{JV}, \langid{KN}, \langid{SU}, \langid{SW}, \langid{YO} & NLI   & sent &  -    & \XSolidBrush    & 3709  & \Checkmark   \\ \midrule
\meta & \langid{ES}, \langid{EN} & det, NLI   & tok, sent &  all  & \Checkmark    & 13320  & \Checkmark   \\ \bottomrule

\end{tabular}
\end{table}

It can be argued that these shortcomings may produce misleading results and conclusions regarding the ability of language models to understand metaphorical language. In order to bridge this gap, we present the first parallel corpus including: (i) metaphorical annotations for metaphor detection at the token level covering nouns, adverbs, adjectives, and verbs, and (ii) metaphor interpretation annotations grounded in the NLI task for both \langid{ES} and \langid{EN} languages. In addition, this resource allows for large-scale cross-lingual and multilingual experimentation on both tasks (detection and interpretation) with data sourced from naturally occurring language, which was not synthetically generated via lexical substitution. Table \ref{tab:dataset_comparison} summarizes the main features of the most popular datasets for metaphor processing reviewed in this section. 

\section{\meta Corpus}

This section describes the development of \meta, a parallel dataset in \langid{ES} and \langid{EN} newly annotated with metaphorical annotations for detection and interpretation. In the following subsections, we describe the collection of the dataset (Subsection \ref{sec:meta_desc}), the methodology employed to annotate metaphor for each task and language (Subsection \ref{sec:annotation_process}) and, finally, details of the resulting \meta dataset (Subsection \ref{sec:result_dataset}). We finish by discussing the inter-annotator agreement in Section \ref{sec:agreement} and properties of our dataset to study cross-lingual transfer of metaphorical expressions (Section \ref{sec:corpus_analysis}).

\subsection{Description} \label{sec:meta_desc}
% Meta description and motivation
\meta is a compilation of \textbf{\xnli} \cite{conneau-etal-2018-xnli} and \textbf{\esxnli} \cite{artetxe-etal-2020-translation}. We decided to use these data sources since we evaluate metaphor interpretation in the NLI framework. This enables both annotations at the token level for detection and sentence level for interpretation. In addition, it consists of parallel text, from which we select data in \langid{ES} and \langid{EN}. Moreover, the combination of \xnli and \esxnli constitutes a dataset of larger size compared to commonly available resources for metaphor processing, and with natural language utterances and spontaneous usage of metaphors. The distribution of \meta is detailed in Table \ref{tab:meta_distribution}.

% XNLI description
\xnli dataset is a cross-lingual evaluation set for MultiNLI \cite{williams-etal-2018-broad}. It contains parallel data with original text in \langid{EN} subsequently human-translated to other 14 languages, among which \langid{EN} and \langid{ES} were selected for manual annotation of metaphor. It includes a total of 7500 premise-hypothesis pairs from 10 text genres, namely, 830 premises and 2490 hypotheses from \xnlidev, and 1670 premises and 5010 hypotheses from \xnlitest. 
\esxnli comprises a total of 2490 pairs, as in \xnlidev, from 5 different genres. In contrast to \xnli, sentences were originally collected in \langid{ES} and then human-translated to \langid{EN}. The direction of translation (\langid{EN} > \langid{ES} in \xnli and \langid{ES} > \langid{EN} in \esxnli) is an interesting feature to explore the cross-lingual transfer of metaphorical expressions. 

\xnli and \esxnli share the same collection methodology: a set of premise sentences was crawled from various sources in \langid{EN} and \langid{ES}, respectively. Afterwards, crowd-source workers were asked to generate three hypotheses for each premise, one for each label. Both corpora are balanced in terms of inference tags and text domains.

\begin{table*}[h!]
\caption{Number of sentences from each source dataset composing \meta.}
\label{tab:meta_distribution}
\begin{tabular}{lrrrr}
\toprule
              & \textbf{\xnlidev} & \textbf{\xnlitest} & \textbf{\esxnli} & \textbf{\meta} \\
\midrule
Premises      & 830      & 1670      & 830     & 3330  \\
Hypotheses    & 2490     & 5010      & 2490    & 9990  \\
\midrule
Total         & 3320     & 6680      & 3320    & 13320 \\
\bottomrule
\end{tabular}

\end{table*}

\subsection{Annotation Process} \label{sec:annotation_process}

The methodology to label \meta varies across tasks and languages. Although four annotators were involved in the whole process, manual annotation was mostly performed by a linguist native in Spanish with proficient knowledge of English and an expert in metaphor processing annotation.

\subsubsection{Detection}
Annotations for this task are performed at the token level, since we approach metaphor detection as a sequence labeling task. We extract premise and hypothesis sentences and annotate them separately. Therefore, we do not take into account the premise as context to annotate its corresponding hypotheses and vice versa. Due to this split, the total number of labeled sentences is 13320. With respect to the type of metaphors, we consider as candidates the tokens belonging to a semantically significant part of speech (POS): nouns, verbs, adjectives, and adverbs. 

We adopt the MIPVU guidelines \cite{steenetal2010} throughout the annotation process, either for manual revision or in automatic predictions, since models used were trained on data labeled accordingly to this procedure. The guidelines used can be summarised in the four main steps enumerated in \cite{steenetal2010}:

\begin{enumerate}
    \item Read the entire text/discourse to establish a general understanding of the meaning.
    \item Determine the lexical units in the text/discourse.
    \item \begin{enumerate}
        \item For each lexical unit in the text, establish its meaning in context, that is, how it applies to an entity, relation, or attribute in the situation evoked by the text (contextual meaning). Take into account what comes before and after the lexical unit.
        \item For each lexical unit, determine if it has a more basic contemporary meaning in other contexts than the one in the given context. For our purposes, basic meanings tend to be:
                \begin{itemize}
                    \item More concrete; what they evoke is easier to imagine, see, hear, feel, smell, and taste.
                    \item Related to bodily action.
                    \item More precise (as opposed to vague).
                    \item Historically older.
                    Basic meanings are not necessarily the most frequent meanings of the lexical unit.
                \end{itemize}
        \item If the lexical unit has a more basic current–contemporary meaning in other contexts than the given context, decide whether the contextual meaning contrasts with the basic meaning but can be understood in comparison with it.
    \end{enumerate}
    \item If yes, mark the lexical unit as metaphorical.
    
\end{enumerate}

\begin{figure}
        \centering
        \includegraphics[width=0.9\linewidth]{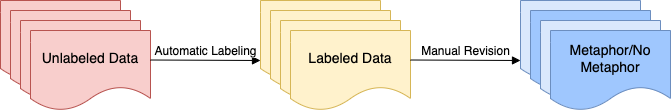}
        \caption{Annotation process for \metaes. Premises and hypotheses were automatically labeled. Subsequently, we manually reviewed all automatic predictions. The resulting dataset contains sentences with metaphors labeled at the token level.}
        \label{fig:spanish_det_annotation}
\end{figure}

\paragraph{\textbf{Spanish}} The annotation process in this language comprehends two phases, as depicted in Figure \ref{fig:spanish_det_annotation}. First, we automatically label \metaes by leveraging \mdeberta \cite{He2021DeBERTaV3ID} fine-tuned for metaphor detection in \langid{ES} with \cometa. We chose \mdeberta since it is the multilingual model that achieved the highest F1 score in the experimental setup of \citet{sanchez-bayona-agerri-2022-leveraging}. This choice reduces the heavy workload and time investment that a manual annotation from scratch would require. Afterwards, we manually inspect and correct the predictions for the whole dataset. From the first phase of automatic labeling, 748 tokens were predicted as metaphorical usages of words in the premises and 724 in the hypotheses. After a complete and manual revision of the 13320 sentences, 74 tokens were removed, and 481 undetected metaphors in the premises were added. Regarding the hypotheses, we deleted 118 false positives and labeled 533 false negatives.

The main sources of ambiguity in \langid{ES} emerge from \textbf{multi-word expressions (MWE)} and \textbf{polysemy}.
The main issue when labeling a \textbf{MWE} is to decide whether to treat the MWE as a single lexical unit, a fixed expression where the meaning of each word is not transparent anymore, or whether it should be regarded as a collocation. Collocations are expressions where the constituent words tend to co-occur with high frequency but are not fixed, since each of their elements can be replaced by others with similar meaning.

\enumsentence{Dijo que era hora de \textit{entrar} en pánico (lit. ``They said it was time to enter into panic'').} \label{ex:mwe_entrar}

For instance, in Example \ref{ex:mwe_entrar}, the expression \textit{entrar en pánico} could be initially considered as an MWE with a single lexical unit, therefore, the three tokens would be labeled as metaphorical. However, this expression specifically means ``to panic'', in which the verb \textit{entrar} (lit. ``to enter'') holds metaphorical meaning, since ``panic'' is not a physical place you can get into. In this case, we do not think of the expression as a fixed MWE, since the verb is also used metaphorically with other terms that are not places, like \textit{entrar en cólera}  (``to get angry'', lit. ``to enter into wrath'') or \textit{entrar en calor}  (``to feel hot'', lit. ``to enter into heat''). In all of these expressions, the verb is conveying the sense of starting to feel the noun it complements, as if by entering a place, it transformed our sensitivity. This association of concepts arises from understanding emotions or sensations as physical locations. 

Regarding \textbf{polysemy}, the existence of multiple and very nuanced senses associated with the same token can lead to confusion. It can be challenging to determine whether the basic meaning of the lexical unit is currently and generally known and used by native speakers or if they directly associate the lexical unit with the figurative meaning, not identifying the basic meaning at all. 

\enumsentence{[...] ha mostrado su \textit{apoyo} a la candidatura para ser sede [...] (lit. ``They showed their support to the candidacy to be head office'').} \label{ex:polysemy_apoyo}

For instance, in Example \ref{ex:polysemy_apoyo} we label \textit{apoyo} as metaphorical, since the most basic meaning of the verb \textit{apoyar} in the dictionary defines it as ``to make something rest upon another thing''. In this sentence, the contextual meaning refers to someone in favour of someone else's goal. Figuratively, the goal can be understood in terms of a physical object so heavy that it requires more than one anchor point to distribute its weight. Doubts in this kind of examples arise from the fact that the figurative sense may be more frequent than its basic one, thus speakers might not identify the metaphorical meaning as such. We use the \emph{Diccionario de la Real Academia Española} (DRAE) as a tool to help us clarify these ambiguous cases. Nonetheless, metaphor identification remains a subjective task.

\paragraph{\textbf{English}}

\begin{figure}[h]
    \centering
    \includegraphics[width=1\linewidth]{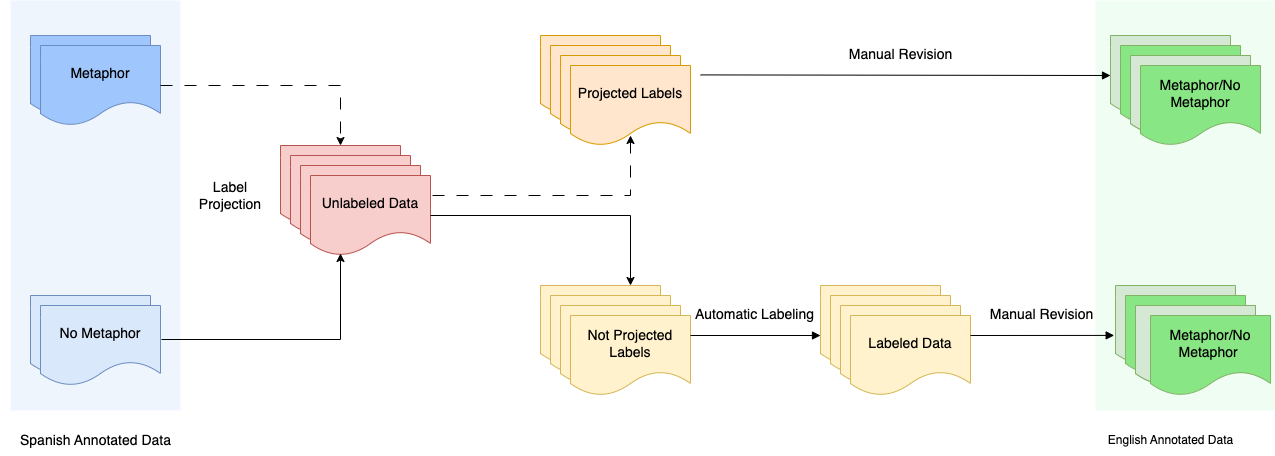}
    \caption{Annotation process for \metaen. We took \langid{ES} annotations as source to project metaphor labels onto their \langid{EN} counterpart sentences. Sentences without labels in \langid{ES} were not transferred to \langid{EN}, therefore, we automatically labeled this subset and manually reviewed it. The resulting dataset is the combination of sentences with projected and predicted labels at the token level.}
    \label{fig:english_det_annotation}
\end{figure}

We generate \metaen annotations semi-automatically and based on \langid{ES} annotations, since our purpose is to publish the parallel resource with annotations for metaphor detection and interpretation. The annotation process consists of four phases, as depicted in Figure \ref{fig:english_det_annotation}. 

% Projections and manual revision
The first step involves the projection of \langid{ES} labels onto \langid{EN} sentences using Easy-Label-Projection \cite{garcia-ferrero-etal-2022-model}, a tool developed for cross-lingual sequence labeling that makes use of word alignments and data- and model-transfer to project the labels from a source language (\langid{ES}) to an untagged target language (\langid{EN}). This technique is appropriate when the labeled entity is certainly appearing in both source and target sentences. However, metaphors present in a source sentence are not necessarily in its translation. It depends on a series of multiple factors, namely the type of translation, the knowledge of the translator, whether it is human-translated, or socio-cultural knowledge, among others. The projected labels are manually revised.

% Automatic labeling and manual revision
The next step involves the automatic annotation (to minimize manual effort) using language models of the sentences without any metaphor annotation after executing the Spanish to English projection step (87\% of the total). We applied \xlmroberta \cite{xmlroberta} fine-tuned on the \vuam dataset (the best multilingual model in \langid{EN} in the experiments by \citet{sanchez-bayona-agerri-2022-leveraging}). We manually reviewed the automatic predictions from XLM-RoBERTa to correct errors and undetected metaphorical expressions following the MIPVU procedure. The total number of metaphorical tokens is reported in the next Section \ref{sec:result_dataset}.

% Issue gained in translation
Some issues emerged during the whole annotation process. Firstly, the projections phase entails that for a metaphorical expression to be annotated in \langid{EN}, it has to have been annotated in \langid{ES} in advance. Hence, metaphors in \langid{EN} sentences that were not expressed figuratively in their \langid{ES} counterparts will not be spotted, as in Example \ref{ex:false_negative_english}. In the \langid{ES} sentence, there is no metaphorical expression annotated as such. However, in the \langid{EN} version, the adjective \textit{heavy} is holding metaphorical meaning, as a synonym of ``demanding'', which is expressed literally in \langid{ES} with the adjective \textit{exigentes}. In this type of case, the lack of annotation in the source language implies that no label will be projected onto the target sentence, missing a metaphorical instance. These examples give an account of how translation and the singularities of metaphors according to languages may affect the annotation process of this task. We provide an analysis of these cases in Section \ref{sec:corpus_analysis}.

\eenumsentence{\item{A los usuarios más exigentes se les debería cobrar más.} 
\item{The \textit{heavy} users should be charged the most.} }\label{ex:false_negative_english}

% Issue lost in translation and pronouns
Another issue worth discussing is that of false positives: every \langid{ES} sentence with one or more labeled metaphors will transfer those tags to the \langid{EN} sentence, regardless of whether the translated tokens have metaphorical meaning or not. To address this, we manually reviewed all sentences with a projected metaphorical tag. This step removed metaphors that were ``lost in translation'' and adjusted some spans projected to the target language, e.g. some annotated verbs in \langid{ES} were projected in \langid{EN} to the subject pronoun and verb, since some \langid{ES} verb forms are synthetic and include the person information in a single morpheme. Therefore, we eliminated the label from the pronoun and kept only that of the verb. In Example \ref{ex:labeled_subject_english} we can see how the verb \textit{peleaban} (lit. ``they fought'') labeled as metaphorical in the \langid{ES} sentence (Example \ref{ex:labeled_subject_spanish}) was projected onto the subject and verb in \langid{EN} (Example \ref{ex:labeled_subject_projection}). Example \ref{ex:labeled_subject_final} represents the definitive version of the annotations after manual revision.  

\eenumsentence{\item \textit{Peleaban} por lo ricos que eran los directores ejecutivos. \label{ex:labeled_subject_spanish}
             \item \textit{\st{They} fought} about how rich CEOs were. \label{ex:labeled_subject_projection} 
            \item They \textit{fought} about how rich CEOs were. \label{ex:labeled_subject_final}
            }\label{ex:labeled_subject_english}

% Issues automatic labelling - ambiguity and difference of annotation criterion
Regarding the subset of sentences that were automatically labeled by \xlmroberta, some concerns emerged with respect to annotations present in the \vuam dataset during the phase of manual revision. Especially when it comes to phrasal verbs and abstract and polysemous terms that are lexicalised. As illustrated by examples \ref{ex:lexicalised_make} and \ref{ex:prasal_get_rid}, which are labeled as metaphorical in the training set \vuam, we observed a tendency to overannotate verbs that compose phrasal verbs as metaphorical, such as \textit{get}, \textit{look}, \textit{made}, or \textit{go}. In most cases, these verbs appear in a context where they do not add any strong semantic information due to their lexicalisation. Hence, we unmarked these instances that were predicted as metaphorical. Following this line of thought, we also discarded abstract and vague terms that are common-places of spontaneous discourse, like the word \textit{thing} in Example \ref{ex:abstract_thing}, since it can refer to any kind of entity, either concrete or abstract and might not be directly matched to a more broadly used basic meaning.

\enumsentence{His lack of humbug about political \textit{balance} has always \textit{made} him more honest than all the employees [...].}\label{ex:lexicalised_make} 
\enumsentence{Take what you want and leave the rest, your mother'll \textit{get} rid of it .} \label{ex:prasal_get_rid}
\enumsentence{One \textit{thing} always \textit{linked} to another \textit{thing}. } \label{ex:abstract_thing}

\subsubsection{Interpretation}

Similarly to other works discussed in Section \ref{sec:related_work}, we frame metaphor interpretation within the task of NLI \cite{agerri2008metaphor, mohler2013applying, chakrabarty-etal-2021-figurative, stowe-etal-2022-impli, kabra-etal-2023-multi}. Our approach aims at evaluating the capability of LMs to identify the inferential relationship when there is a metaphorical expression either in the premise or, hypothesis, or in both sentences. To do so, we labeled premise-hypothesis pairs with metaphorical expressions where the understanding of the figurative expression is crucial to determine the inference label. We summarised this annotation process in the following steps:

\begin{enumerate}
    \item Read the premise sentence and determine its general meaning.
    \item Identify potential metaphorical expressions according to metaphor detection guidelines.
    \item Repeat the previous two instructions with the hypothesis sentence. 
    \item Establish the inference relation between premise and hypothesis if not previously labeled.
    \item If there is any metaphorical expression either in the premise or the hypothesis:
    \begin{enumerate}
        \item Is it required to understand the literal meaning of the metaphorical expression to label the inference relationship between premise and hypothesis?
        \begin{itemize}
            \item Yes: mark the pair.
            \item No: mark the pair as a non-relevant case.
        \end{itemize}
    \end{enumerate}
    \item Repeat the process with non-relevant cases until clarification. Otherwise, if they are intrinsically ambiguous or lack context to either identify the metaphors or determine if they are relevant to the inference, discard the pair.    
\end{enumerate}

In Example \ref{ex:metaphor_affects}, we encounter the metaphor \textit{saltar} (lit. ``to skip'') with the meaning of omitting some information and not the literal sense of physically jumping. In the hypothesis, the sentence refers to the intention of telling the other interlocutor all the information, comprehensively, without ignoring any part, which contradicts the overall meaning of the premise. The understanding of this metaphorical expression, thus, is required to infer that they contradict each other.

\enumsentence{\textbf{Premise}: Hay tanto que se puede decir sobre eso, que sencillamente me voy a \textit{saltar} eso. (lit. ``There is so much you can say about that, that I am simply going to \emph{skip} that'').\\
\textbf{Hypothesis}: ¡Quiero contarte todo lo que sé sobre eso! (lit. ``I want to tell you everything I know about it!''). \\
\textbf{Inference label}: contradiction} \label{ex:metaphor_affects}

\enumsentence{\textbf{Premise}: No hay necesidad de \textit{hurgar} en ese tema, a menos que quieras asegurarte de que nos hundimos.	(lit. ``There is no need to \emph{rummage} in that topic, unless you want to make sure we will sink'').\\
\textbf{Hypothesis:} Hay una forma de que se hundan. (lit. ``There is one way to make them sink''). \\
\textbf{Inference label}: entailment } \label{ex:borderline}

Non-relevant cases comprehend premise-hypothesis pairs where the understanding of the literal sense of the metaphor is not essential to establish a relationship of entailment, contradiction, or neutrality. As we can see in Example \ref{ex:borderline}, the metaphorical expression \textit{hurgar} (lit. ``to rummage'') is used with a sense of exploring an unpleasant topic, while the literal meaning implies physically digging into an inner space. The entailment in this example is inferred from the likelihood of a sinking, not focusing on the willingness to talk about the mentioned topic. Thus, the interpretation of the metaphorical expression is not relevant to extracting the inference relation.

We set aside the latter group as non-relevant cases, since we are not certain as to which extent the role of metaphor is significant in these occurrences. As a result, we discriminate three classes: a) pairs with metaphors that are relevant to the inference relationship, b) pairs with metaphors that are not relevant to the inference, and c) pairs without metaphors. 

Annotations were manually developed on \langid{ES} text and were transferred to \langid{EN}. Hence, we provide \metaen annotations as a silver standard for further refinement. Data about the number of samples and labels will be resumed in the following Subsection \ref{sec:result_dataset}.

\subsection{Resulting Dataset}\label{sec:result_dataset}
The outcome of this annotation process is \meta, the first parallel dataset with labels for the tasks of metaphor detection and interpretation via NLI in \langid{ES} and \langid{EN}.

\paragraph{\textbf{Detection}}
The parallel data for this task comprises a total of 13320 sentences annotated at the token level, since we approached metaphor detection as a sequence labeling task, following the criteria of the cited previous work in Section \ref{sec:related_work}. 

With respect to \textbf{Spanish} annotations, there is a total of \textbf{1155 metaphorical tokens in premises and 1139 in hypotheses}. Out of the 13320 sentences, 1873 contain at least one metaphorical expression, which constitutes 14\% of the whole dataset. More details in Table \ref{tab:spanish_annotations_detection} according to the various data sources and splits. The premises present a higher proportion of metaphors. This might be due to the fact that premises are longer than hypotheses \citep{conneau-etal-2018-xnli}. In addition, the premises were collected from naturally occurring utterances, while hypotheses were generated by crowd-sourced workers. Therefore, hypothesis sentences might tend to be shorter and simpler.

\begin{table}[h!]
\caption{\textbf{Metaphor annotations for detection in \langid{ES}}: Number of tokens annotated as metaphors and sentences that contain at least one metaphorical expression in each source dataset. Percentages in the \textit{Total} row are averages across datasets.}
\label{tab:spanish_annotations_detection}
\footnotesize
\begin{tabular}{ccrrr|rrr}
\toprule
\multirow{2}{*}{} & \multirow{2}{*}{} & \multicolumn{3}{c|}{\textbf{Tokens}} & \multicolumn{3}{c}{\textbf{Sentences}} \\
                  &                   & Met & Total & Met \% & Met & Total & Met \% \\ 
\midrule
\multirow{3}{*}{\xnlidev} 
  & All           & 552  & 40493 & 1.36 & 455 & 3320 & 13.73 \\
  & Hyp           & 276  & 24347 & 1.13 & 252 & 2490 & 10.16 \\
  & Prem          & 276  & 16146 & 1.71 & 203 & 830  & 24.46 \\
\midrule
\multirow{3}{*}{\xnlitest} 
  & All           & 1027 & 81511 & 1.26 & 864 & 6680 & 12.96 \\
  & Hyp           & 551  & 48703 & 1.13 & 508 & 5010 & 10.18 \\
  & Prem          & 476  & 32808 & 1.45 & 356 & 1670 & 21.32 \\
\midrule
\multirow{3}{*}{\esxnli} 
  & All           & 715  & 42635 & 1.68 & 554 & 3320 & 16.69 \\
  & Hyp           & 312  & 23261 & 1.34 & 278 & 2490 & 11.16 \\
  & Prem          & 403  & 19374 & 2.08 & 276 & 830  & 33.25 \\
\midrule
\textbf{Total} 
  &               & 2294 & 164639 & 1.39 & 1873 & 13320 & 14.06 \\
\bottomrule
\end{tabular}

\end{table}

Regarding \textbf{English} annotations, we can observe a similar trend to that of \langid{ES} in Table \ref{tab:english_annotations_detection}. A total of 3330 tokens were labeled as metaphors, and 2736 sentences contain at least one metaphorical instance out of the total 13320 sentences. Premises show a higher metaphor ratio than hypotheses, as in \langid{ES} annotations. Additionally, we observe a larger amount of labeled metaphors in \langid{EN}, which we attribute to the different annotation processes specified in Subsection \ref{sec:annotation_process}. More specifically, given that \vuam contains a significantly higher amount of labeled metaphorical expressions, the MLM fine-tuned with this dataset predicted many ambiguous metaphors, which we partially removed in a manual revision. These discrepancies are not only noticeable in the annotations but also in the experimental results. Moreover, this gives an account of how guidelines for metaphor identification labeling are open to discussion and clarification, due to the subjective and nuanced nature of this cognitive-linguistic phenomenon.

\begin{table}[h!]
\caption{\textbf{Metaphor annotations for detection in \langid{EN}}: Number of tokens annotated as metaphors and sentences that contain at least one metaphorical expression in each source dataset.}
\label{tab:english_annotations_detection}
\footnotesize
\begin{tabular}{ccrrr|rrr}
\toprule
\multirow{2}{*}{} & \multirow{2}{*}{} & \multicolumn{3}{c|}{\textbf{Tokens}} & \multicolumn{3}{c}{\textbf{Sentences}} \\
                  &                   & Met & Total & Met \% & Met & Total & Met \% \\ 
\midrule
\multirow{3}{*}{XNLI\textsubscript{dev}} 
  & All           & 826  & 38369 & 2.15 & 678 & 3320 & 20.42 \\
  & Hyp           & 442  & 23271 & 1.90 & 388 & 2490 & 15.58 \\
  & Prem          & 384  & 15098 & 2.54 & 290 & 830  & 34.94 \\
\midrule
\multirow{3}{*}{XNLI\textsubscript{test}} 
  & All           & 1543 & 76990 & 2.00 & 1301 & 6680 & 19.48 \\
  & Hyp           & 864  & 46578 & 1.85 & 769  & 5010 & 15.35 \\
  & Prem          & 679  & 30412 & 2.23 & 532  & 1670 & 31.86 \\
\midrule
\multirow{3}{*}{esXNLI} 
  & All           & 961  & 41462 & 2.32 & 757 & 3320 & 22.80 \\
  & Hyp           & 411  & 22540 & 1.82 & 361 & 2490 & 14.50 \\
  & Prem          & 550  & 18922 & 2.91 & 396 & 830  & 47.71 \\
\midrule
\textbf{Total} 
  &               & 3330 & 156821 & 2.12 & 2736 & 13320 & 20.54 \\
\bottomrule
\end{tabular}

\end{table}

% Please add the following required packages to your document preamble:
% \usepackage{multirow}
\begin{table}[h!]
\caption{\textbf{Metaphor annotations for interpretation}: number of premise-hypothesis pairs according to metaphor occurrence. \textit{Met}: pairs where the understanding of a metaphorical expression is required to label the inference relationship. Non-relevant (\textit{Non -rel}) cases comprise pairs with metaphors that are not essential to extract the inference relationship. \textit{No met}: pairs without metaphors.}
\label{tab:metaphor_interpretation_annotations}
\footnotesize
\begin{tabular}{clrcccrcr}
\toprule
                            &      & Met & Met \% & Non-rel & Non-rel\% & No met & No met\% & Total \\ \toprule
\multirow{4}{*}{\xnlidev}  & All  & 289      & 11.61     & 449        & 18.03       & 1752        & 70.36        & 2490  \\
                            & Ent  & 101      & 12.17     & 136        & 16.39       & 593         & 71.45        & 830   \\
                            & Neu  & 96       & 11.57     & 153        & 18.43       & 581         & 70.00        & 830   \\
                            & Cont & 92       & 11.08     & 160        & 19.28       & 578         & 69.64        & 830   \\ \midrule
\multirow{4}{*}{\xnlitest} & All  & 580      & 11.58     & 758        & 15.13       & 3672        & 73.27        & 5010  \\
                            & Ent  & 190      & 11.38     & 236        & 14.13       & 1244        & 74.49        & 1670  \\
                            & Neu  & 188      & 11.26     & 270        & 16.17       & 1212        & 72.57        & 1670  \\
                            & Cont & 202      & 12.10     & 252        & 15.09       & 1216        & 72.81        & 1670  \\ \midrule
\multirow{4}{*}{\esxnli}     & All  & 378      & 15.18     & 528        & 21.20       & 1584        & 63.61        & 2490  \\
                            & Ent  & 134      & 16.14     & 168        & 20.24       & 528         & 63.61        & 830   \\
                            & Neu  & 116      & 13.98     & 183        & 22.05       & 531         & 63.98        & 830   \\
                            & Cont & 128      & 15.42     & 177        & 21.33       & 525         & 63.25        & 830   \\ \midrule
Total                       &   & 1247     & 12.48     & 1735       & 17.37       & 7008       & 70.15        & 9990 \\ \bottomrule
\end{tabular}

\end{table} 

\paragraph{\textbf{Interpretation}}
Annotations for this task were developed at the premise-hypothesis level. As shown in Table \ref{tab:metaphor_interpretation_annotations}, the average percentage of pairs with metaphors relevant to the inference relationship is 12\%. This figure remains steady throughout each source dataset and inference labels. \esxnli shows a higher number of metaphor occurrences that might be caused by the difference in text domains and sentence characteristics with respect to \xnli data. Regarding non-relevant cases, we do not use them in this work to be able to focus on establishing whether metaphor presence impacts models' performance. We keep the same sample distribution for both languages in every experiment.

\subsection{Inter-annotator Agreement} \label{sec:agreement}
In a first round of annotation, the main annotator 1 revised the English and Spanish splits as explained in the previous section. In order to calculate inter-annotator agreement (IAA), we selected a subset of 1000 pairs in \langid{ES}. This subset was labeled by annotator 2,  another Spanish native speaker proficient in English and with a linguistics background. Annotator 2 labeled this subset from scratch and according to MIPVU guidelines in the case of detection, and the annotation process specified in Subsection \ref{sec:annotation_process} for interpretation. We computed Cohen's Kappa score \cite{Cohen1960ACO} and obtained 0.74 in premises and 0.77 in hypotheses sentences for detection; for interpretation, the IAA score was 0.64.

We performed a second round of annotation to compare the quality of the annotations semi-automatically generated with full manual labeling. Two hired professional annotators (native Spanish with proficient knowledge of English) manually labelled 6660 instances in English and Spanish (13320 in total). This second iteration resulted in an agreement of 0.452 (Cohen’s kappa between the main original annotator 1 and the new hired annotator 4) and a Fleiss (among all three main annotators) of 0.428 (moderate agreement).

Table \ref{tab:inter_annotator_agreement} presents the inter-annotator agreement scores of the three main annotators for metaphor detection in English and Spanish. The Cohen’s Kappa values reach a fair-moderate agreement (0.373 Cohen’s Kappa) in English, namely, between the new annotations (annotator 3) and the original annotations (semi-automatic) revised by annotator 1. Moreover, IAA between the two manual annotations (annotators 3 and 4) is quite similar, namely, 0.413 (Cohen’s Kappa, moderate agreement). Although for Spanish the IAA scores are slightly higher, they exhibit the same patterns. This shows the consistency of our original annotation with respect to the fully manual one introduced in this second iteration.

Still, metaphor annotation is inherently subjective and context-dependent, and these results reflect that complexity. The moderate kappa values obtained by the second iteration show that while annotators agree on many instances, there remain ambiguous cases that challenge consistent labeling. Furthermore, the differences between the IAA rates obtained between the two sets of annotators (original and the newly hired ones) can be attributed to the longer experience of the original two annotators in collaborating in the manual labeling of metaphor detection. Nonetheless, the obtained IAA scores show more than a moderate agreement and are comparable to those obtained in the annotation of other datasets for metaphor processing \cite{jang-etal-2014-conversational, kesarwani-etal-2017-metaphor, sanchez-montero-etal-2025-disagreement}.

\begin{table}[ht]
\caption{Inter-annotator agreement for metaphor detection annotation using exact labels. Cohen's $\kappa$ is reported for each annotator pair; Fleiss' $\kappa$ reflects overall agreement among all three main annotators.}
\label{tab:inter_annotator_agreement}
\begin{tabular}{lcccc}
\toprule
\textbf{Language} & \textbf{Cohen's $\kappa$ (1–3)} & \textbf{Cohen's $\kappa$ (1–4)} & \textbf{Cohen's $\kappa$ (3–4)} & \textbf{Fleiss' $\kappa$} \\
\midrule
\langid{EN} & 0.373 & 0.369 & 0.413 & 0.374 \\
\langid{ES} & 0.404 & 0.452 & 0.438 & 0.428 \\
\bottomrule
\end{tabular}

\end{table}

\subsection{Lexical Properties and Cross-Lingual Transfer} \label{sec:corpus_analysis}
\paragraph{Lexical distribution}

\begin{table}[h!]
\caption{Comparative lexical properties across splits and languages.}
\label{tab:lexical_distribution}
\begin{tabular}{lrrrrrrrrrrrr}
\toprule
 & \multicolumn{3}{c}{\langid{ES}} & \multicolumn{3}{c}{\langid{EN}} \\
\cmidrule(lr){2-4} \cmidrule(lr){5-7}
\textbf{} & Train& Dev& Test& Train& Dev& Test\\
\midrule
Metaphor tokens& 1053& 474& 767& 1527& 697& 1106\\
Once-occurring metaphor tokens& 722& 387& 596& 962& 536& 802\\
Sentences ($\geq$2 metaphors)  & 126& 80& 116& 198& 110& 161\\
 Avg sentence length (tokens)& 10.94& 13.28& 14.57& 10.46& 12.64&13.81\\
 \bottomrule
\end{tabular}

\end{table}

We conducted a corpus analysis with a focus on lexical properties and the distribution of metaphorical expressions to better understand the dataset’s characteristics. Table \ref{tab:lexical_distribution} presents the number of metaphorical tokens in the training, development, and test sets for each language, along with counts of metaphorical tokens that occur only once, the number of sentences containing multiple metaphors, and the average sentence length. The analysis reveals that English contains a higher number of metaphorical tokens overall, whereas Spanish sentences tend to be longer on average.

\begin{figure}[h!]
    \centering
    \begin{minipage}[t]{0.48\linewidth}
        \centering
        \includegraphics[width=\linewidth]{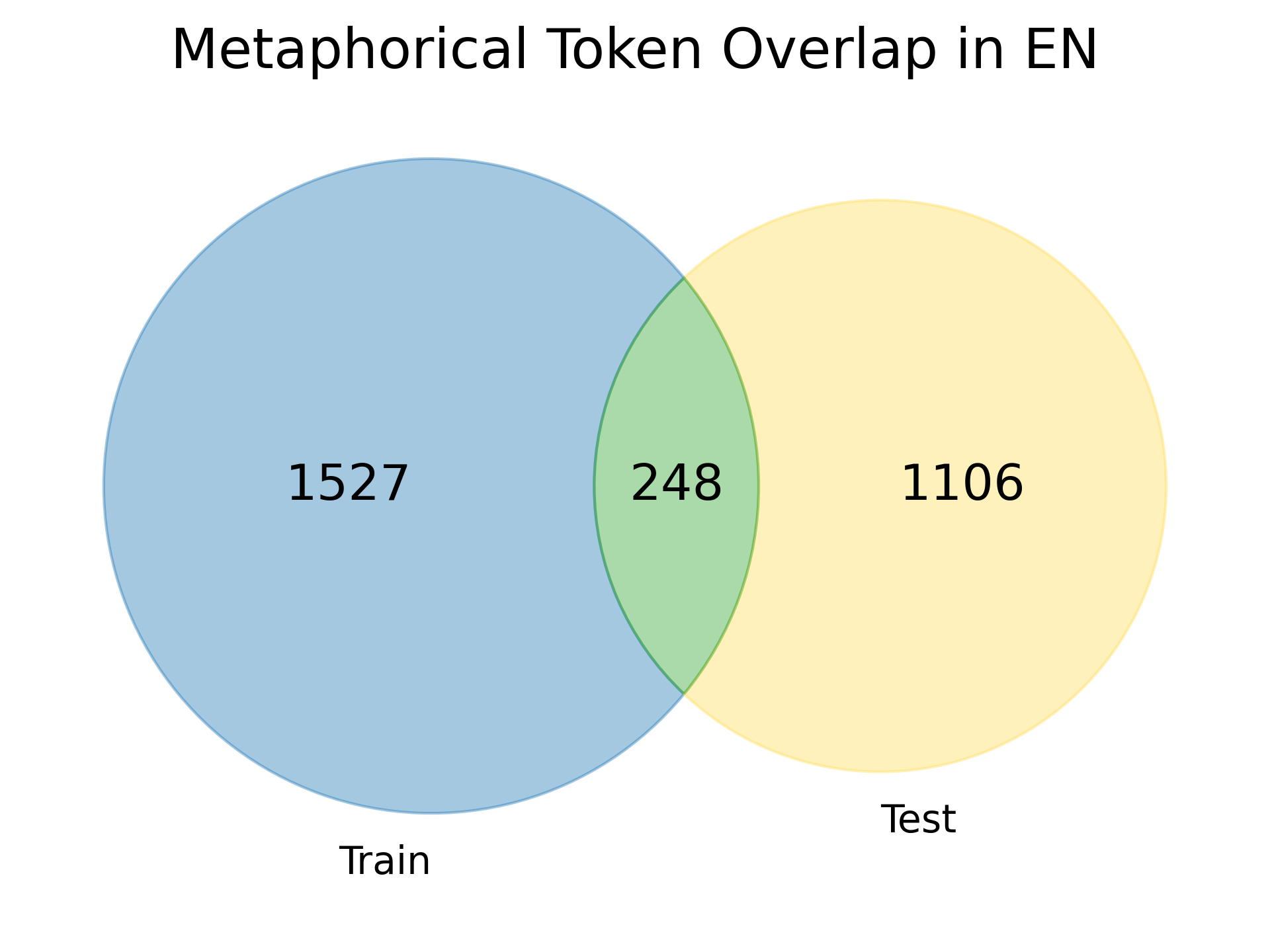}
        \caption{Metaphorical tokens overlap by exact match between train and test in English.}
        \label{fig:met_overlap-english}
    \end{minipage}
    \hfill
    \begin{minipage}[t]{0.48\linewidth}
        \centering
        \includegraphics[width=\linewidth]{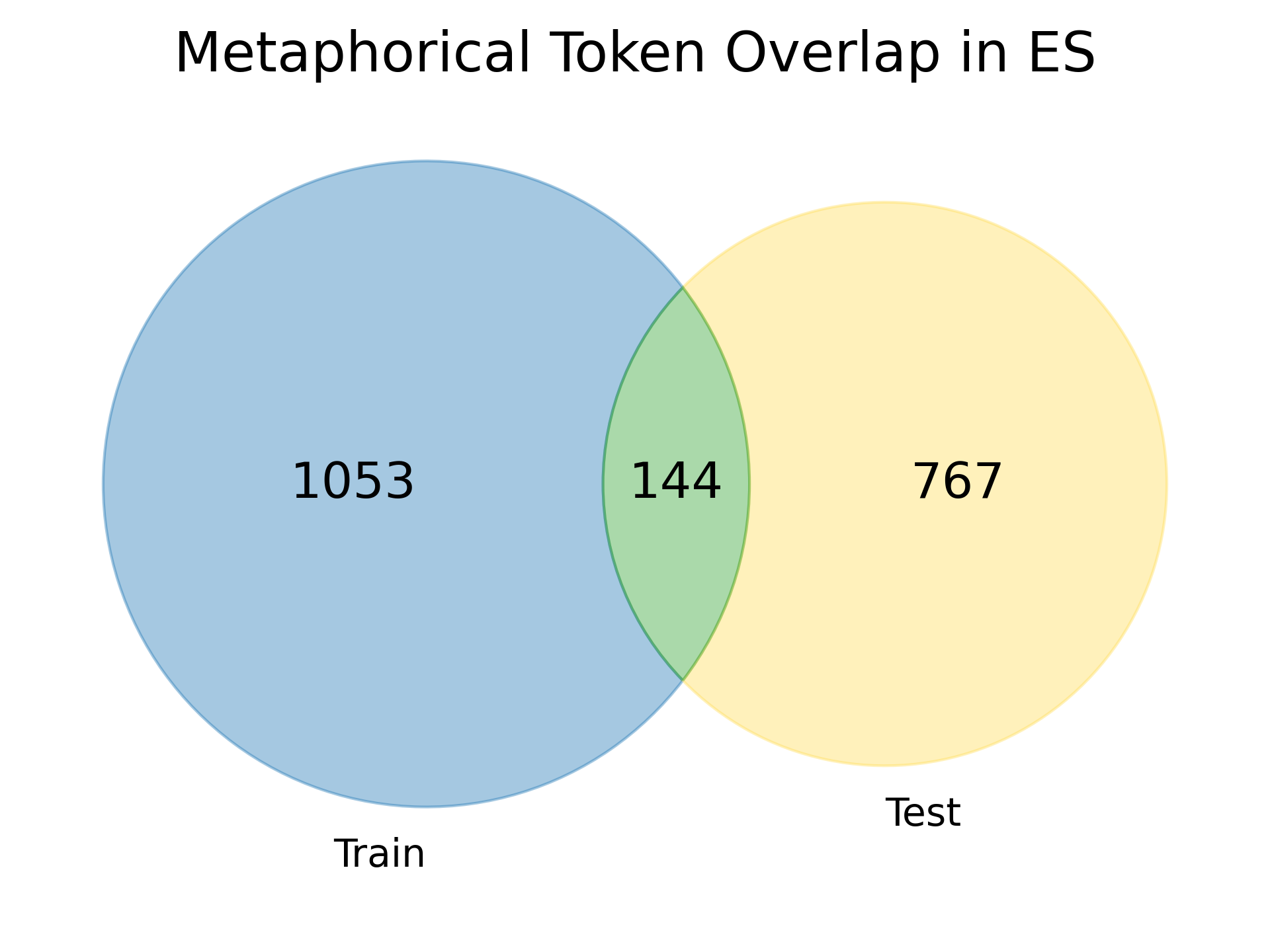}
        \caption{Metaphorical tokens overlap by exact match between train and test in Spanish.}
        \label{fig:met_overlap-spanish}
    \end{minipage}
\end{figure}

\begin{figure}[h!]
    \centering
    \begin{minipage}[t]{0.5\textwidth}
        \centering
        \includegraphics[width=\linewidth]{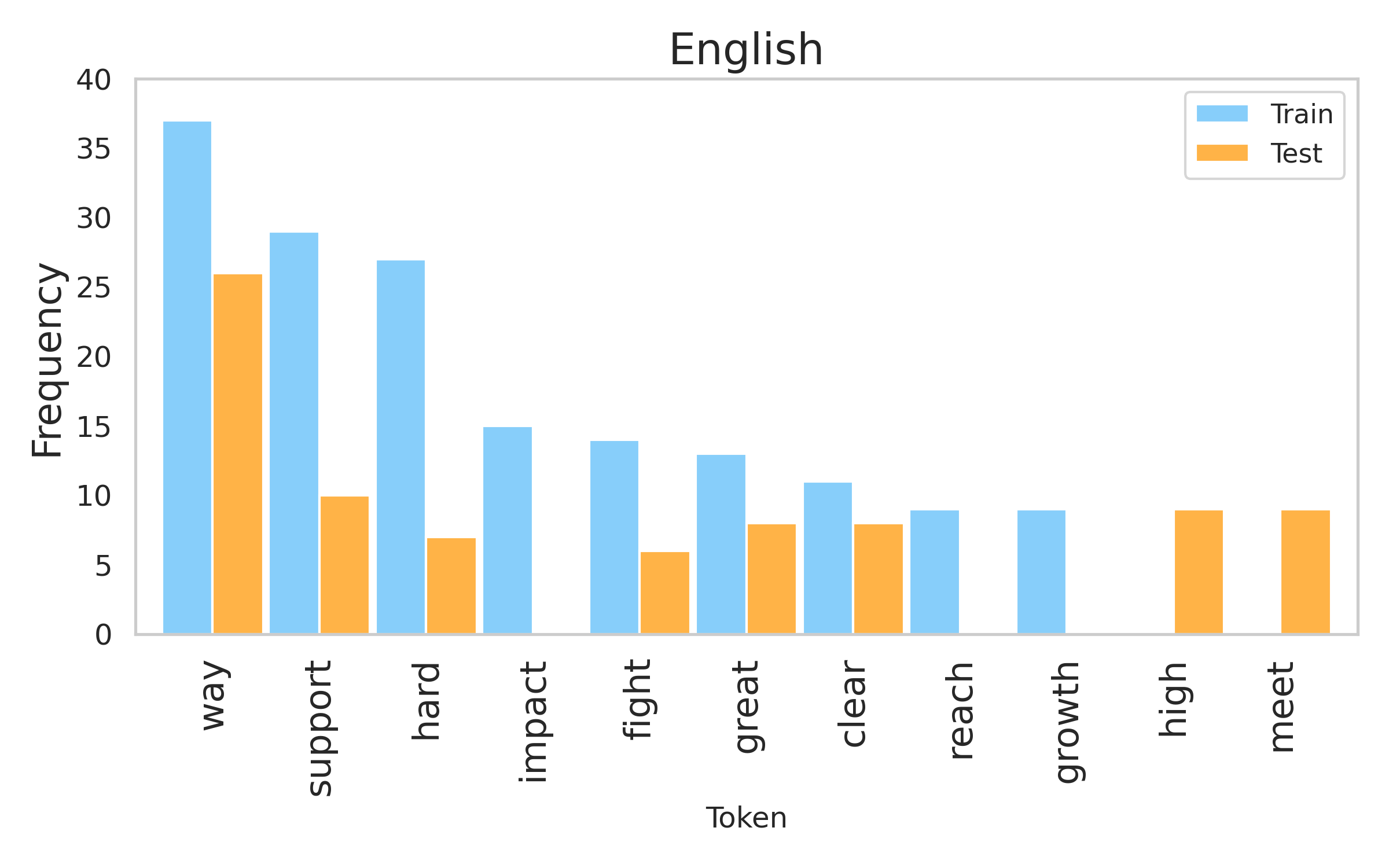}
        \caption{Most frequent metaphorical tokens in train and test in English.}
        \label{fig:freq_metaphor_english}
    \end{minipage}%
    \hfill
    \begin{minipage}[t]{0.5\textwidth}
        \centering
        \includegraphics[width=\linewidth]{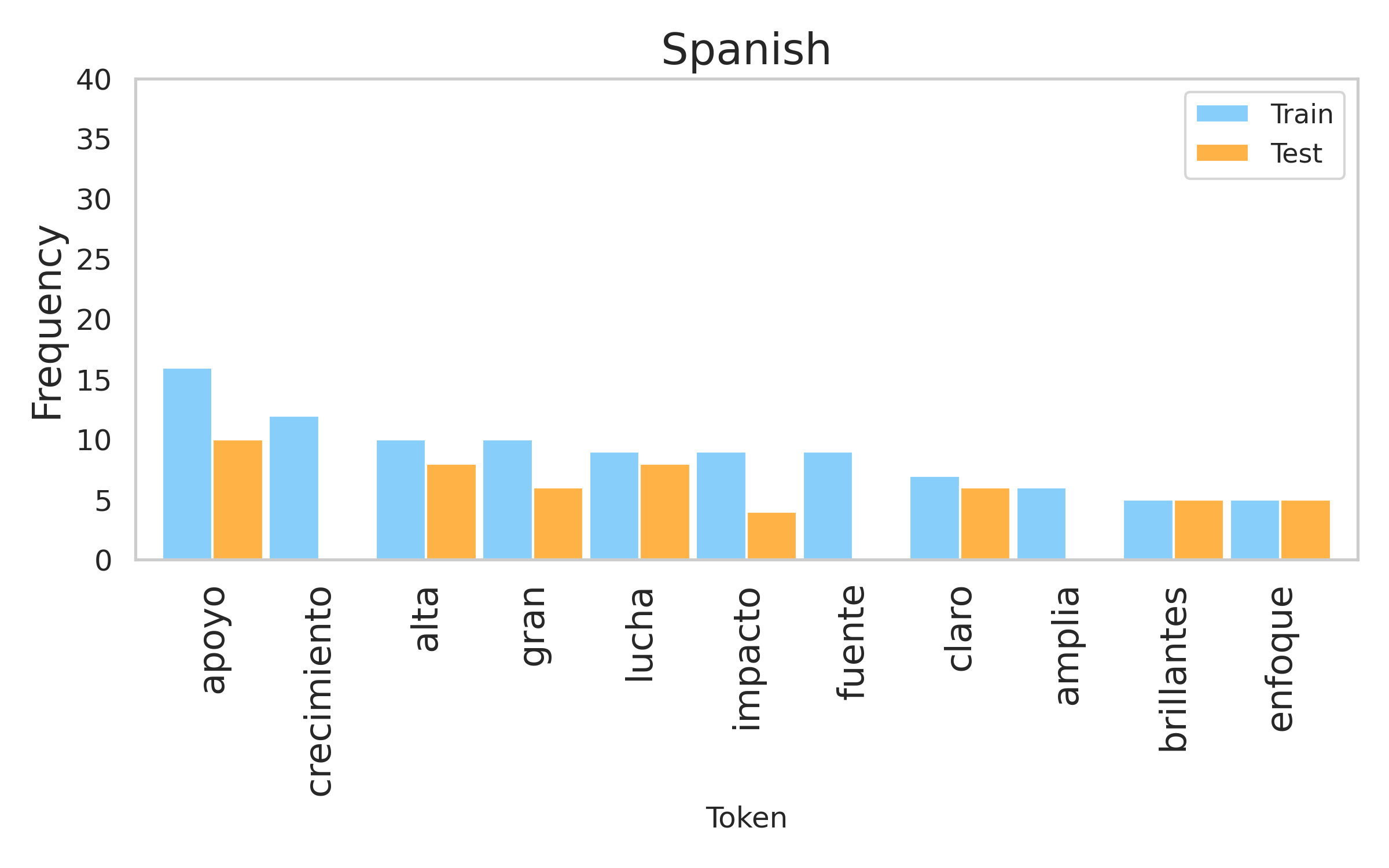}
        \caption{Most frequent metaphorical tokens in train and test in Spanish.}
        \label{fig:freq_metaphors_spanish}
    \end{minipage}
\end{figure}

To better comprehend models' performance and their generalization abilities, we examined the token overlap between training and test sets using exact match, as shown in Figures \ref{fig:met_overlap-english} and \ref{fig:met_overlap-spanish}. In English, overlapping tokens account for 22.42\% of the test set, while in Spanish, the overlap is 18.77\% in the test set. These figures indicate a relatively low metaphor overlap. A more informative analysis could be conducted at the conceptual level in the future.

Examples of this overlap are illustrated in Figures \ref{fig:freq_metaphor_english} and \ref{fig:freq_metaphors_spanish}, which display the most frequent metaphorical tokens within each partition. Many of the most frequent metaphor tokens are shared between the training and test sets, and in some cases, across both languages. These high-frequency tokens often correspond to conventional metaphorical expressions, which may influence the results of metaphor detection and interpretation tasks.

\paragraph{\textbf{Cross-lingual transfer}}
We analysed metaphor transfer across languages, with quantitative results presented in Table \ref{tab:metaphor_transfer}. As discussed in Section \ref{sec:annotation_process}, the higher number of metaphor annotations in English can be attributed to the semi-automatic annotation process applied to this language. In particular, the model used to identify metaphors in English was fine-tuned on the VUAM corpus, which contains a significantly higher proportion of labeled metaphorical tokens \cite{sanchez-bayona-agerri-2022-leveraging}.

After manual inspection of individual cases, shown in Table \ref{tab:metaphor_transfer_examples}, we found that one of the primary factors contributing to metaphor transference is the translation process. For example, instances 1, 2, 3, and 7 illustrate cases where metaphors present in the original version were expressed literally in the translation. These changes appear to be introduced during the translation process, carried out by humans in these datasets. In instance 4, for example, the metaphor ``barriers'' is translated as ``obstáculos'' in Spanish, which is a difficulty, but not a physical object that hinders advancement, like a ``barrier'' (``barrera'' in Spanish). In contrast, instance 5 shows a case where the metaphorical expression is maintained in both languages.

The transference of some metaphorical expressions also stems from language-specific expressions or collocations. In instance 7, for example, the English verb ``to spend'' from the MONEY domain is used for the TIME domain. However, in Spanish, this verb does not naturally occur to express this meaning. Another language-specific phenomenon is illustrated in instance 6: while both languages use a metaphorically equivalent expression, English relies on phrasal verbs that are MWEs, whereas Spanish typically encodes the same meaning in a single token.

\begin{table}[ht]
\caption{\textbf{Cross-lingual Transfer: } Total count of metaphorical tokens in each language (\textit{ES/EN Met} columns) after the complete annotation process. The last two columns refer to tokens ``lost in translation'', either labeled in \langid{ES} but not expressed metaphorically in the \langid{EN} sentence -> \langid{ES} \Checkmark \langid{EN} \XSolidBrush, or labeled in \langid{EN} but not in \langid{ES}. -> \langid{ES} \XSolidBrush \langid{EN} \Checkmark.} \label{tab:metaphor_transfer}
\footnotesize
\begin{tabular}{llrrrr}
\toprule
&& \textbf{\langid{ES} Met} & \textbf{\langid{EN} Met} & \textbf{\langid{ES} \Checkmark \langid{EN} \XSolidBrush} & \textbf{\langid{ES} \XSolidBrush \langid{EN} \Checkmark} \\
\midrule
\multirow{2}{*}{\xnlidev}    &Hyp& 276  & 442  & 9   & 175 \\
   &Prem& 276  & 384  & 18  & 126 \\
\multirow{2}{*}{\xnlitest}  &Hyp& 551  & 864  & 7   & 320 \\
  &Prem& 476  & 679  & 30  & 233 \\
\multirow{2}{*}{\esxnli}       &Hyp& 312  & 411  & 35  & 134 \\
     &Prem& 403  & 550  & 34  & 181 \\
\midrule
Train           & - & 1053 & 1527 & 49  & 523 \\
Dev               &- & 474  & 697  & 34  & 257 \\
Test              & - & 767  & 1106 & 50  & 389 \\
\bottomrule
\end{tabular}
\end{table}
 
\begin{table}[ht]
\caption{\textbf{Cross-lingual Transfer}: Examples of metaphor shifts between Spanish and English sentences in the dataset.} \label{tab:metaphor_transfer_examples}
\footnotesize
\renewcommand{\arraystretch}{1.3} % More spacing between rows
\resizebox{\textwidth}{!}{
\begin{tabular}{>{\bfseries}r p{4.5cm} p{4.5cm} >{\centering\arraybackslash}p{1.8cm} >{\centering\arraybackslash}p{1.8cm} >{\raggedright\arraybackslash}p{1.5cm}}
\toprule
\# & \textbf{Spanish} & \textbf{English} & \textbf{Met. Tokens (ES)} & \textbf{Met. Tokens (EN)} & \textbf{Source} \\ \midrule
1& Britney y su familia han estado desconectando en Malibú. & Britney and her family have been relaxing in Malibu. & desconectando & --- & \esxnli \\
2& El mundo es en gran parte fruto de la ciencia. & The world is in large part a product of science. & fruto & --- & \esxnli \\
3& La COPE y Esradio tienen una relación tormentosa. & COPE and Esradio have a difficult relationship. & tormentosa & --- & \esxnli \\
4& Dos años después, aún existían obstáculos para compartir la información. & Information sharing barriers were still in place after two years. & --- & barriers & \xnlidev \\
5& El director no se creía que las barreras de intercambio de información se tuvieran que eliminar completamente. & The Director did not believe that information sharing barriers should be entirely removed. & barreras & barriers & \xnlidev \\
6& Tienes que desarrollar cuidadosamente una organización de CIO o se derrumbará. & You have to develop a CIO organization carefully or it will fall apart. & derrumbará & fall, apart & \xnlitest \\
7& Pasó tiempo en los Estados Unidos. & He spent time in the United States. & --- & spent & \xnlitest \\
\bottomrule
\end{tabular}}

\end{table}

\section{Experimental setup} \label{sec:experiments}

In this section, we present the settings for the experiments designed to test the capabilities of multilingual MLMs (or encoder-only models) and LLMs (or decoder-only models). We evaluated metaphor detection with MLMs in cross-domain, cross-lingual, and multilingual scenarios. To assess LLMs, we additionally fine-tuned a series of models in monolingual and multilingual settings. We limited the experimental scenarios with LLMs due to the computational costs. Concerning metaphor interpretation, we also experiment with their ability to perform NLI when the correct inference requires understanding metaphorical language. We used the same encoders as for detection. In the case of decoders, we added larger models for inference with zero-shot and chain-of-thought (CoT) prompts, available in Appendix Table \ref{tab:eval-prompts}.

With the aim of developing cross-lingual experiments, we chose the multilingual language encoder-only models and checkpoints that obtained best results in \citet{sanchez-bayona-agerri-2022-leveraging}: \mdeberta (base) and \xlmroberta (large) \cite{xmlroberta}, the multilingual versions of DeBERTa \cite{He2021DeBERTaV3ID} and RoBERTa \cite{liu2019roberta}, respectively, both for detection and interpretation. For decoder-only models, we fine-tuned \llamasm \cite{dubey2024llama3herdmodels}, \qwensm \cite{qwen2.5} and \gemma \cite{gemmateam2024gemmaopenmodelsbased} for metaphor detection experiments, since models with a larger number of parameters are computationally expensive. Fine-tuning LLMs was based on the method by \citet{garcia-ferrero-etal-2024-medmt5}, which enables fine-tuning and inference on LLMs for sequence labeling tasks with constrained decoding. In the case of metaphor interpretation, we evaluated \llamalg, \qwenlg, \gpt through in-context learning in order to test LLMs understanding capabilities. Experiments with open-weight models were performed via the HuggingFace implementations. For \gpt \footnote{\url{https://platform.openai.com/}} we used the API.

\paragraph{\textbf{Detection}}

Taking advantage of previously available resources and \meta, we conducted a series of experiments to evaluate and fine-tune MLMs and LLMs. 
The configuration of cross-domain experiments is specified in \citet{sanchez-bayona-agerri-2022-leveraging}. For the other three setups with encoders, monolingual, multilingual, and zero-shot cross-lingual experiments, we performed hyperparameter tuning through grid search with batch size [8, 16, 32], linear decay [0.1, 0.01], learning rate (in the [1e-5-5e-5] interval), sequence length of 128 and epochs from 4 to 10. A warm-up of 6\% is specified. The results of the hyperparameter tuning showed that after 4 epochs, development loss started to increase, so the results reported here are obtained with 4 epochs, batch size of 8, weight decay of 0.1 and learning rate of 5e-5. 

For the fine-tuning of decoders in mono- and multilingual setups, we also performed grid search over the development set and the following parameter space: learning rate [2e-4, 2e-5, 2e-6, 2e-7],  max sequence length 512, batch size [8, 16, 32], epochs [5, 10, 30] and a fixed seed. We report the best results obtained with 30 epochs, 8 batch size and 2e-5 learning rate.

\begin{itemize}
    \item \textbf{Cross-domain:} the aim of this set of experiments is to evaluate the performance of MLMs fine-tuned with \cometa and \vuam datasets on \metaes and \metaen, respectively, since each dataset contains texts from different domains. Specifically, text sources from \xnli cover Face-To-Face, Telephone, Government, 9/11, Letters, Oxford University Press, Slate, Verbatim, and Government and Fiction. \esxnli is a compilation of texts from 5 sources: a newspaper, an economic forum, a celebrity magazine, a literature blog, and a consumer magazine. The motivation is to explore the impact of text features and genres on the performance, as well as annotation criteria \cite{probing-aghazadeh-etal-2022-metaphors, multi-multifigurative}. To do so, we chose the models with the best performance from monolingual experiments developed in \citet{sanchez-bayona-agerri-2022-leveraging}. We conducted the evaluation on various data splits: within each source dataset, \xnlidev, \xnlitest and \esxnli, we evaluated premises and hypotheses separated and combined. This is due to the dissimilarities and the unequal distribution of metaphorical expressions between premises and hypothesis sentences mentioned in Subsection \ref{sec:result_dataset}.
    
    \item \textbf{Monolingual:} this scenario comprises the fine-tuning and evaluation of MLMs and LLMs on \metaes and \metaen separately. To accomplish this task, we split \meta into train, development, and test sets (0.6-0.2-0.2). We equally distributed the data to ensure each partition is balanced in terms of source datasets, sentences, and metaphor occurrence. Data statistics are detailed in the Appendix. These partitions will be used as well in subsequent multilingual and zero-shot cross-lingual scenarios. In addition to fine-tuning and evaluation on \metaes and \metaen, we evaluated each trained monolingual model with the test sets of \cometa and \vuam, following the same reasoning as in cross-domain experiments. 
    
    \item \textbf{Multilingual:} The purpose of these experiments is to explore whether MLMs and LLMs benefit from being trained on data in multiple languages. In this case, we combined \metaes and \metaen train splits to fine-tune the models. Subsequently, we evaluated the trained models on each language test set, in order to analyse the impact on the performance for each language. The data splits used correspond to those from monolingual experiments.
    \item \textbf{Zero-shot cross-lingual:} in this scenario, we explore to what extent MLMs are able to generalize knowledge and metaphor transfer between these two languages in question. Therefore, we fine-tune the models with \meta data in one language and evaluate them on the test set of the other. Data partitions used for these experiments are the same as in monolingual and multilingual scenarios.
\end{itemize}

\paragraph{\textbf{Interpretation}}
We carried out two sets of experiments to evaluate metaphor interpretation within the NLI task with encoder-only models. We performed hyperparameter tuning through grid search, with the same range of parameters specified on the task of metaphor detection. We report the best results obtained on the development set, after 4 epochs, batch size of 8, learning rate of 1e-5, weight decay of 0.1 and 512 sequence length. 

To assess decoder-only models' performance on the NLI task with sentences with and without metaphors, we performed inference over the specified models with the following parameters and Hugging Face and vLLM \cite{kwon2023efficient} implementations: max new tokens=5, temperature=0.3. Models were prompted to answer with one of the three NLI relations [entailment, natural, contradiction]. To do so, we designed two prompts available in Appendix Table \ref{tab:eval-prompts}: one with no examples (zero-shot) and another one with longer context and one example for each label (chain-of-thought (CoT)). We only report the results for the first set of experiments, due to the computational cost of fine-tuning models of this size.

The purpose of the first experiments is to examine whether the presence of metaphorical expressions in premise-hypothesis pairs impacts the performance of models in the NLI task. To that end, we fine-tuned the MLMs for the task with the MultiNLI dataset. Then, we evaluated them with \meta. Among each source datasets, we discriminated pairs with metaphors relevant to the inference relationship from those without metaphorical expressions. We developed the evaluation on each of these subsets and for each language separately, e.g., one evaluation on \xnlidev metaphor set and another on \xnlidev without metaphors set, in \langid{EN} and then in \langid{ES}. 
The goal of the second set of experiments is to analyse the effect on models' performance of not being ``exposed'' to instances with metaphorical expressions during training. With this aim in mind, we extracted pairs with and without metaphors from \meta train, dev, and set splits. In the first scenario, we fine-tuned the models only with pairs from the train set that did not contain any metaphorical expressions. In the second scenario, we mixed pairs with and without metaphors and fine-tuned the models as well. In both cases, we evaluated on the test sets with and without metaphors for each language.

\section{Results} \label{sec:results}

\paragraph{\textbf{Detection}}
In addition to F1 score, we computed \textbf{in-vocabulary} (\textit{Inv}) and \textbf{out-of-vocabulary} (\textit{Oov}) F1 scores to assess the impact of the labels seen from training data during the learning process of the models. Precision and Recall metrics are reported in Appendix C. For the in-vocabulary evaluation, we calculated the F1 score using only the predicted metaphorical tokens that also appeared in the training data labeled as metaphors. In contrast, the out-of-vocabulary evaluation was computed on predicted metaphorical tokens that were not labeled as metaphors in the training set. We used \emph{exact match} to extract the overlapping tokens. We included this evaluation in all experiments, but for the zero-shot cross-lingual setup, since the data of the train and test sets are in different languages, the match of the same metaphorical token in both partitions is highly unlikely. 

Although the purpose of our experiments is not to beat state-of-the-art results but to evaluate the performance of MLMs and LLMs on the task from a cross-lingual approach, we added two \textbf{relevant baselines}: on one hand, the system \textbf{BasicBERT} \cite{li-etal-2023-metaphor-meanings-model}, which obtained \textbf{73.3} F1 score; on the other, the result of \textbf{DeBERTa} reported in \cite{sanchez-bayona-agerri-2022-leveraging}, with \textbf{73.79} F1. We selected these results for comparison purposes, since they both were evaluated on the VUAM-2020 version of the dataset in \langid{EN} used in the Shared Task 2020 \cite{leong-etal-2020-report} and in the same experimental setup we propose. 

% Please add the following required packages to your document preamble:
% \usepackage{multirow}
\begin{table}[h!]
\caption{\textbf{Cross-domain metaphor detection: } F1 scores from evaluation of \meta on models trained with other metaphor detection datasets of different textual domain. \textit{Inv} stands for ``in-vocabulary'' evaluation, which only takes into account metaphor tokens seen in training and \textit{Oov}, from ``out-of-vocabulary'' evaluation, which only takes into account predicted metaphor tokens not seen during the training process. Best model performance in bold.}
\label{tab:xnli_det_crossdomain}
\footnotesize
\begin{tabular}{ll|ccc|ccc}
\toprule
\multirow{3}{*}{\textbf{Train (\langid{ES})}} & \multirow{3}{*}{\textbf{Test (\langid{ES})}} & \multicolumn{3}{c|}{\multirow{2}{*}{\textbf{\mdeberta}}}      & \multicolumn{3}{c}{\multirow{2}{*}{\textbf{\xlmroberta}}} \\
&   & \multicolumn{3}{c|}{} & \multicolumn{3}{c}{} \\
                        &                  & F1 & Inv   & Oov  & F1   & Inv   & Oov  \\ \midrule
\multirow{11}{*}{\cometa} & \esxnli all      & \textbf{64.94} & 69.03 & 60.41 & 60.52 &                                 63.75     & 56.78    \\
                        & \esxnli hyp      & \textbf{58.78} & 66.88 & 48.07 & 55.58 & 
                        58.00 & 52.13    \\
                        & \esxnli prem     & \textbf{69.74} & 70.97 & 68.56 & 64.22 & 68.82  & 59.65  \\ \cmidrule{2-8} 
                        & \xnlidev all   & \textbf{62.61} & 64.41 & 60.98 & 58.11 & 61.71  & 54.70    \\
                        & \xnlidev hyp   & \textbf{63.49} & 66.06 & 61.36 & 59.49 & 66.67  & 53.53    \\
                        & \xnlidev prem  & \textbf{61.67} & 62.83 & 60.53 & 56.62 & 56.89  & 56.34    \\ \cmidrule{2-8} 
                        & \xnlitest all  & \textbf{60.11} & 61.08 & 59.19 & 58.46 & 60.38 & 56.58    \\
                        & \xnlitest hyp  & \textbf{57.89} & 58.09 & 57.68 & 55.63 & 57.83 & 53.36    \\
                        & \xnlitest prem & \textbf{62.70} & 64.78 & 60.83 & 61.63 & 63.40 & 60.00    \\ \cmidrule{2-8}
                        & average & \textbf{62.44} & 64.90 & 59.73 & 58.92 & 61.94 & 55.90 \\\cmidrule{2-8}
                        & \cometa & 67.46 & - & - & 67.44 & - & - \\ 
                        \bottomrule

\end{tabular}

\begin{tabular}{ll|ccc|ccc}
\multirow{3}{*}{\textbf{Train (\langid{EN})}} & \multirow{3}{*}{\textbf{Test (\langid{EN})}} & \multicolumn{3}{c|}{\multirow{2}{*}{\textbf{DeBERTa}}} & \multicolumn{3}{c}{\multirow{2}{*}{\textbf{\xlmroberta}}} \\
                        &                  & \multicolumn{3}{c|}{} & \multicolumn{3}{c}{} \\
                        &                  & F1   & Inv  & Oov  & F1   & Inv   & Oov  \\ \midrule
\multirow{11}{*}{\vuam}   & \esxnli all      & \textbf{38.02} & 36.89 & 41.75 & 35.23  & 34.36  & 38.41    \\
                        & \esxnli hyp      & \textbf{32.63} & 32.73 & 32.30 & 30.60  & 30.67  & 30.74    \\
                        & \esxnli prem     & \textbf{43.46} & 41.32 & 50.09 & 39.88  & 38.29  & 45.02    \\ \cmidrule{2-8} 
                        & \xnlidev all   & \textbf{37.52} & 34.57 & 45.66 & 34.67  & 32.32  & 40.86    \\
                        & \xnlidev hyp   & \textbf{35.69} & 33.47 & 43.03 & 33.02  & 31.26  & 38.64    \\
                        & \xnlidev prem  & \textbf{40.09} & 36.23 & 48.72 & 36.83  & 33.83  & 43.20    \\ \cmidrule{2-8} 
                        & \xnlitest all  & \textbf{34.37} & 33.17 & 38.45 & 32.09  & 31.13  & 35.57     \\
                        & \xnlitest hyp  & \textbf{31.70} & 31.21 & 34.64 & 30.18  & 29.88  & 32.71    \\ 
                        & \xnlitest prem & \textbf{38.13} & 36.16 & 42.90 & 34.75  & 33.02  & 38.80 \\ \cmidrule{2-8}
                        & average & 36.85 & 35.08 & 41.95 & 34.14 & 32.75 & 38.22 \\ \cmidrule{2-8}
                        & \vuam & 73.79 & - & - & 72.11  & -  & - \\ 
    \bottomrule
\end{tabular}
\end{table}

\textbf{Cross-domain:} We evaluated models trained on \cometa and \vuam datasets with \xnli and \esxnli in \langid{ES} and \langid{EN}, respectively. From results reported in Table \ref{tab:xnli_det_crossdomain}, we can observe that in all cases \mdeberta outperforms \xlmroberta for \langid{ES}. In \langid{EN} the best result is obtained by DeBERTa, which also outperforms \xlmroberta in all scenarios. In all datasets, except for \xnlidev in \langid{ES}, premise sentences achieve better results than hypotheses and the combination of both. This goes in line with annotation statistics in Tables \ref{tab:spanish_annotations_detection} and \ref{tab:english_annotations_detection}, which show that premises contain a greater ratio of metaphors per sentence than hypotheses in both languages. In \langid{ES}, in-vocabulary evaluation outperforms the general F1 score while out-of-vocabulary evaluation results decrease. The small distance in points of this cross-domain evaluation in \langid{ES} exhibits stability of the models when it comes to predicting metaphors and coherence in annotations between both datasets, despite the difference of text domains. 

Nevertheless, \langid{EN} results do not demonstrate such consistency, as out-of-vocabulary obtains higher results than the overall F1 score. These discrepancies are also reflected in a significant drop in performance with respect to \langid{ES} experiments and in-domain evaluation with \vuam. Thus, the high recall scores from the Appendix Table \ref{tab:appendix_xnli_det_crossdomain} show that the model tends to predict many metaphors; however, low precision scores indicate that a small amount of these predictions are correct. Previous work showed that even though MIPVU guidelines (used to annotate VUAM) were applied for the manual annotation of the CoMeta dataset, the guidelines allow for a level of subjectivity in their interpretation, which means that the proportion of metaphorical tokens was around 6\% of the total for VUAM, while in CoMeta it is around 2\% \cite{sanchez-bayona-agerri-2022-leveraging}. 

Regarding \meta, the Spanish data was fully revised manually, while, for English, we used a semi-automatic approach (described in Section \ref{sec:annotation_process}) based on first deploying encoders fine-tuned on English VUAM data and then a manual revision of the automatic annotations. This means that, during the manual revision phase, the English dataset already contained preliminary labels automatically annotated with models fine-tuned on VUAM. As a consequence, when comparing the English with the manually annotated Spanish split,  and just as it happened between CoMeta and VUAM, the English split contains a higher number of annotated metaphorical tokens. Thus, while the differences are not as large as in the case of CoMeta and VUAM (where domain differences need to be factored in), the English Meta4XNLI contains around 6\% more metaphorical tokens than the Spanish split. 

\textbf{Monolingual:} Results from this set of experiments are specified in Tables \ref{tab:xnli_det_finetune} and \ref{tab:xnli_det_finetune_llms}. After fine-tuning and evaluating the models with \metaes and \metaen, the highest overall F1 score is obtained by \mdeberta in \langid{ES} for encoders, and \llamasm for decoders. On the other hand, \xlmroberta and \gemma achieve the best performance in \langid{EN} but still lower than in \langid{ES}. 
\begin{table*}[h]
\caption{
\textbf{Monolingual metaphor detection encoder-only models}: F1 score results from model fine-tuning with \meta (M4X) and evaluation with its test set language and \vuam (\langid{EN}) and \cometa (\langid{ES}) test sets, for each corresponding language. The score is an average of results from 5 random runs, standard deviation next to F1 scores. Best model performance in bold.
}
\label{tab:xnli_det_finetune}
\footnotesize
\begin{tabular}{cc|ccc|ccc} 
\toprule
    \multicolumn{2}{c}{} & \multicolumn{3}{c}{\textbf{\mdeberta}} & \multicolumn{3}{c}{\textbf{\xlmroberta}} \\
    \textbf{Train} & \textbf{Test} & F1 & Inv & Oov & F1 & Inv & Oov  \\
     \midrule
     \multirow{2}{*}{M4X\scriptsize{\langid{ES}}}  & 
            M4X\scriptsize{\langid{ES}} &  \textbf{67.17} \tiny{$\pm$ 0.72} & 71.89 \tiny{$\pm$ 1.48} & 60.41 \tiny{$\pm$ 2.41} & 66.84 \tiny{$\pm$ 0.87} &  70.78 \tiny{$\pm$ 0.83} &  61.24 \tiny{$\pm$ 1.34} \\
            & CoMeta & 55.62 \tiny{$\pm$ 0.61} & 61.95 \tiny{$\pm$ 1.82} & 47.70 \tiny{$\pm$ 2.90} & \textbf{56.24} \tiny{$\pm$ 1.28} & 62.48 \tiny{$\pm$ 1.94} & 48.60 \tiny{$\pm$ 1.07} \\
            \midrule
    
     \multirow{2}{*}{M4X\scriptsize{\langid{EN}}}  & 
            M4X\scriptsize{\langid{EN}} &  58.02 \tiny{$\pm$ 0.95} & 60.94 \tiny{$\pm$ 0.78} & 50.86 \tiny{$\pm$ 2.15} & \textbf{59.24}  \tiny{$\pm$ 1.00} &  61.04 \tiny{$\pm$ 1.21} & 55.30 \tiny{$\pm$ 1.08} \\
            & VUAM & 29.14 \tiny{$\pm$ 1.63} & 26.45 \tiny{$\pm$ 1.73} & 36.86 \tiny{$\pm$ 2.07} & \textbf{32.42} \tiny{$\pm$ 2.68} & 29.14 \tiny{$\pm$ 2.32} & 41.55 \tiny{$\pm$ 3.03}  \\
    \bottomrule
\end{tabular}

\end{table*}
 
\begin{table*}[h]
\footnotesize

\caption{
\textbf{Monolingual metaphor detection decoder-only models}: F1 score results from model fine-tuning with \meta (M4X) and evaluation with its test set language and \vuam (\langid{EN}) and \cometa (\langid{ES}) test sets, for each corresponding language. Best model performance for each test set in bold.
} \label{tab:xnli_det_finetune_llms} 
\begin{tabular}{cc|ccc|ccc|ccc} 
\toprule
    \multicolumn{2}{c}{} & \multicolumn{3}{c}{\textbf{\llamasm}} & \multicolumn{3}{c}{\textbf{\qwensm}} &  \multicolumn{3}{c}{\textbf{\gemma}}\\
    \textbf{Train} & \textbf{Test} & F1 & Inv & Oov & F1 & Inv & Oov   & F1 & Inv &Oov   \\
     \midrule
     \multirow{2}{*}{M4X\scriptsize{\langid{ES}}}  & 
            M4X\scriptsize{\langid{ES}} &  \textbf{59.15}& 63.77& 53.03& 56.76&  62.86&   48.61& 58.06& 63.33&51.17\\
            & CoMeta & \textbf{50.06}& 56.55& 42.74& 44.26& 53.79&  33.23& 45.32& 57.78&36.42\\
            \midrule
    
     \multirow{2}{*}{M4X\scriptsize{\langid{EN}}}  & 
            M4X\scriptsize{\langid{EN}} &  50.41& 53.58& 44.57& 47.66&  51.37&  41.63& \textbf{52.61}& 56.37&45.97\\
            & VUAM & \textbf{26.78}& 22.31& 38.47& 26.44& 21.37&  40.39& 26.72& 22.13&38.94\\
    \bottomrule
\end{tabular}

\end{table*}

In both languages, in-vocabulary evaluation results are higher than the overall ones and out-of-vocabulary results, lower. This is not the case when we use \vuam corpus for testing the model fine-tuned with \metaen. In this setup, similarly to results from cross-domain experiments, performance drastically falls around 30 points of F1 score and out-of-vocabulary results are the highest. The reason might be the mismatches in the annotation process. In \langid{ES}, when evaluating \cometa, we also encounter a decrease in the performance. However, it is a smaller difference of 10 points. This might be caused by the variety of text genres and dissimilarities between sentences from each dataset. Overall, encoder-only models outperform decoder-only models in this task. Despite the decrease of F1 scores, we can observe the same tendencies as with MLM in the results of LLMs in Table \ref{tab:xnli_det_finetune_llms}.
    
\textbf{Multilingual:} In this set of experiments, we assembled \metaen and \metaes to train MLMs and LLMs. The evaluation is conducted for each language separately, and the results are detailed in Table \ref{tab:xnli_det_multilingual}. Encoder-only models outperform decoder-only models in all cases, as in the monolingual experimental setup. Best results in \langid{ES} are obtained by \mdeberta and \llamasm, which are higher than the top result from monolingual experiments. In \langid{EN}, \mdeberta is the model that achieves better performance but is very close to that of \xlmroberta. \llamasm outperform \qwensm and \gemma by a larger margin. The highest F1 score is 8 points lower than that of \langid{ES} but outweighs \langid{EN} monolingual results. This suggests that the combination of parallel multilingual data for training is beneficial for the performance of the models. 
% Required packages:
% \usepackage{booktabs, adjustbox}

\begin{table}[h!]
\caption{\textbf{Multilingual metaphor detection:} Results from models fine-tuned with \metaes + \metaen and evaluated on each language test set individually. Best model performance for each test set in bold. Encoder-only models' score is an average of results from 5 random runs, standard deviation next to F1 scores. Decoder-only models' results with fixed seed.}
\label{tab:xnli_det_multilingual}
\footnotesize
\begin{tabular}{lccc|ccc}
\toprule
& \multicolumn{3}{c}{\textbf{\metaes}} &
\multicolumn{3}{c}{\textbf{\metaen}} \\
& F1 & Inv & Oov & F1 & Inv & Oov \\
\midrule
\textbf{\mdeberta}   & \textbf{68.70} \tiny{$\pm$ 1.27} & 72.63 \tiny{$\pm$ 1.23} & 63.03 \tiny{$\pm$ 1.95} & \textbf{60.80} \tiny{$\pm$ 1.05} & 63.16 \tiny{$\pm$ 1.38} & 55.12 \tiny{$\pm$ 1.08} \\
\textbf{\xlmroberta} & 66.01 \tiny{$\pm$ 0.55} & 70.29 \tiny{$\pm$ 1.18} & 59.84 \tiny{$\pm$ 1.43} & 60.14 \tiny{$\pm$ 1.32} & 61.57 \tiny{$\pm$ 1.14} & 56.79 \tiny{$\pm$ 1.88} \\
\midrule
\textbf{\llamasm}    & \textbf{61.46} & 64.52 & 57.40 & \textbf{55.26} & 57.91 & 50.00 \\
\textbf{\qwensm}     & 58.72 & 63.35 & 52.60 & 52.62 & 56.56 & 46.25 \\
\textbf{\gemma}      & 57.02 & 64.38 & 47.08 & 54.60 & 56.65 & 50.56 \\
\bottomrule
\end{tabular}
\end{table}

\textbf{Zero-shot cross-lingual:} In these experiments, we perform evaluation of \meta in the opposite language to that utilised for training. Results are reported in Table \ref{tab:xnli_det_crosslingual}. \xlmroberta performance exceeds that of \mdeberta in both languages. Nonetheless, F1 score for \langid{EN} is almost 20 points lower than \langid{ES} evaluation results. In addition to the differences in annotation criteria between languages and datasets, another aspect to bear in mind in this scenario could be the number of positive examples present in training sets. As we explained in Section \ref{sec:result_dataset}, \metaen contains a higher number of metaphorical instances, thus models are exposed to a greater variety of examples that can be transferred to \langid{ES}. While a reduced number of instances in \metaes seen during training might hinder the model's generalization ability, as the low recall and high precision scores show in Appendix Table \ref{tab:appendix_xnli_det_crosslingual}.

\begin{table}[h!]
\caption{
\textbf{Zero-shot cross-lingual metaphor detection}: F1 scores of models performance after fine-tuning with\metaes and testing on \metaen, and vice versa. F1 score is an average of 5 random runs, standard deviation next to F1 scores. Best model performance for each evaluation in bold.
}
\label{tab:xnli_det_crosslingual}
\footnotesize
\begin{tabular}{ c  c | c c } 
    \toprule
    \textbf{Train} & \textbf{Test}  & \textbf{\mdeberta} & \textbf{\xlmroberta}  \\
     \midrule
     \metaes & 
            \metaen & 37.78 \scriptsize{$\pm$ 4.35} & \textbf{39.90} \scriptsize{$\pm$ 2.20} \\
            \midrule
    
    \metaen  & 
            \metaes &  56.41 \scriptsize{$\pm$ 3.05} & \textbf{58.94} \scriptsize{$\pm$ 1.32} \\
    \bottomrule        
\end{tabular}
\end{table}

\paragraph{\textbf{Interpretation}}
In the first setup, we fine-tuned the encoder-only models for the NLI task with the MultiNLI dataset \cite{williams-etal-2018-broad}. In the case of decoders, we evaluated the models through in-context learning with different prompts. We conducted the evaluation with two different splits for each source dataset in \meta: pairs with at least one metaphorical expression and pairs lacking metaphors.  
From the accuracy scores reported in Tables \ref{tab:xnli_interpretation_evaluation}, \ref{tab:xnli_interpretation_evaluation_llm}, we can observe certain variability in the results. Encoders' average performance is close to that of decoders like \qwenlg and \gpt. Overall, \gpt is the best decoder-only model for \langid{EN} and \langid{ES} when evaluated with the CoT prompt. 

With respect to encoders, in the majority of cases, \xlmroberta achieves better performance than \mdeberta. We do observe a tendency for higher results on the sets of pairs without metaphors than on the set with metaphors. The exceptions to this current are the results from \xlmroberta for \xnlidev and \xnlitest in \langid{ES}. In these partitions, the subset of pairs with metaphorical expressions obtained better results, although the difference does not even reach one point. 

% Please add the following required packages to your document preamble:
% \usepackage{multirow}
\begin{table}[h!]
\caption{\textbf{Monolingual evaluation of metaphor interpretation with encoder-only models:} Accuracy of models fine-tuned with MultiNLI in the respective language of the test set. Each source dataset conforming \meta was evaluated separately, distinguishing between pairs with and without metaphorical expressions. In bold, best result with respect to metaphor/no metaphor occurrence in each language and source dataset.}
\label{tab:xnli_interpretation_evaluation}
\footnotesize
\begin{tabular}{cccc|cc}
\toprule
                            &                             & \multicolumn{2}{c|}{\textbf{\mdeberta}} & \multicolumn{2}{c}{\textbf{\xlmroberta}} \\
                            &                             & \langid{EN}               & \langid{ES}              & \langid{EN}                & \langid{ES}                \\ \midrule
\multirow{2}{*}{\xnlidev}  & metaphor      & 86.85            & 85.12           & 86.50             & \textbf{86.85}             \\
                            & no metaphor                 & \textbf{87.72}            & \textbf{85.39}           & \textbf{89.15}             & 86.36             \\ \midrule
\multirow{2}{*}{\xnlitest} & metaphor      & 85.02            & 82.27           & 86.40             & \textbf{85.20}             \\
                            & no metaphor  & \textbf{87.74}            & \textbf{85.40}           & \textbf{88.04}             & 84.80             \\ \midrule
\multirow{2}{*}{\esxnli}     & metaphor                    & 74.60            & 78.31           & 74.33             & 77.78             \\
                            & no metaphor                 & \textbf{76.39}            & \textbf{79.17}           & \textbf{76.77}             & \textbf{80.05}   \\
\midrule
 average & & 83.05 & 82.61 & \textbf{83.36} & \textbf{83.51}\\
 \bottomrule
\end{tabular}
\end{table} 
% Please add the following required packages to your document preamble:
% \usepackage{multirow}
\begin{table}[h!]
\caption{\textbf{Monolingual evaluation of metaphor interpretation with decoder-only models:} Accuracy of models evaluated with zero-shot and chain-of-thought prompts for test sets. Each source dataset conforming \meta was evaluated separately, distinguishing between pairs with and without metaphorical expressions. Accuracy scores correspond to the average of 3 runs and standard deviations within the range [0, 0.45]. In bold, best result with respect to metaphor/no metaphor occurrence in each language and source dataset. } \label{tab:xnli_interpretation_evaluation_llm}
\resizebox{\textwidth}{!}{
\begin{tabular}{cccccc|cccc|cccccccc}
\toprule
 & & \multicolumn{4}{c}{\textbf{\llamalg}}& \multicolumn{4}{c}{\textbf{\qwenlg}}& \multicolumn{4}{c}{\textbf{\gpt}}& \\
                   
 & & \multicolumn{2}{c}{zero-shot} & \multicolumn{2}{c|}{chain}& \multicolumn{2}{c}{zero-shot}& \multicolumn{2}{c|}{chain}& \multicolumn{2}{c}{zero-shot} &  \multicolumn{2}{c}{chain}\\
                            &                             & \langid{EN}               & \langid{ES}               & \langid{EN}               &\langid{ES}               & \langid{EN}                & \langid{ES}                  & \langid{EN}               &\langid{ES}               & \langid{EN}                & \langid{ES}                  &  \langid{EN}               &\langid{ES}                \\ \midrule
\multirow{2}{*}{\xnlidev}  & met      & 70.70&  67.82& 77.05&79.12& 82.47&  \textbf{80.85}& 84.31&\textbf{83.73}& 84.26&  \textbf{82.53}&  86.50&\textbf{83.39}\\
                            & no met                 & \textbf{72.72}&  \textbf{69.97}& \textbf{81.14}&\textbf{79.79}& \textbf{85.61}&  80.42& \textbf{86.58}&80.99& \textbf{87.04}&  81.88&  \textbf{86.76}&82.74& \\ \midrule
\multirow{2}{*}{\xnlitest} & met      & 73.39&  \textbf{70.80}& \textbf{82.42}&80.00& 83.10&  \textbf{80.74}& 85.97&80.34& 86.55&  80.52&  \textbf{87.85}&82.07& \\
                            & no met  & \textbf{72.44}&  67.37& 80.55&\textbf{78.71}& \textbf{84.47}&  80.50& \textbf{87.75}&\textbf{81.99}& \textbf{86.72}&  \textbf{82.48}&  87.55&\textbf{82.85}\\ \midrule
\multirow{2}{*}{\esxnli}     & met                   & 72.05&  74.34& 76.02&80.78& 73.10&  84.13& 49.19&83.33& 75.53&  82.67&  76.32&85.32\\
                            & no met                 & \textbf{73.65}&  \textbf{75.86}& \textbf{79.49}&\textbf{83.84}& \textbf{81.82}& \textbf{ 85.42}& \textbf{80.34}&\textbf{85.63}& \textbf{81.50}&  \textbf{85.26}&  \textbf{81.57}&\textbf{86.52}\\
\midrule
average & & 72.56 & 70.55 & 79.89 & 80.32  & 83.28 & 81.71 & 84.72 & 82.57 & 84.61 & 82.54 & \textbf{85.44} & \textbf{83.62} \\
\bottomrule
\end{tabular}}
\end{table}

In Table \ref{tab:xnli_interpretation_evaluation_llm}, we can also observe the same trend: the subsets without metaphors achieve higher scores. These results are more consistent when evaluated in the language of the original dataset (\langid{EN} for \xnli and \langid{ES} for \esxnli), and are more variable in the remaining scenarios. We hypothesize the original language of the dataset might be involved in this performance, since \xnli includes natural utterances of \langid{EN} that were afterwards manually translated to \langid{ES}, thus some artifacts might have been introduced during this process \cite{artetxe-etal-2020-translation}.

The second scenario consisted of fine-tuning the MLMs on two setups: a) with pairs without metaphors and b) pairs with and without metaphors. We performed the evaluation on subsets split by the criterion of metaphor occurrence as well. Results reported in Table \ref{tab:xnli_interpretation_finetune} show \xlmroberta outperforming \mdeberta in all contexts. Both models show best results for the NLI task on the examples without metaphors and the lowest performance with pairs that contain metaphorical expressions. This outcome replicates in both languages and experimental setups a) and b).   
% Please add the following required packages to your document preamble:
% \usepackage{multirow}
\begin{table}[ht]
\caption{\textbf{Monolingual fine-tuning for metaphor interpretation:} Accuracy scores from fine-tuning of models with \meta. On one hand, only instances without metaphors and, on the other, pairs with and without metaphorical expressions. Evaluation is performed on test sets that also discriminate pairs according to metaphor presence in the corresponding language used for training. Results are an average of the accuracy scores from 5 random runs, standard deviation next to accuracy scores. In bold, best result with respect to metaphor/no metaphor occurrence in each language. }
\label{tab:xnli_interpretation_finetune}
\footnotesize
\begin{tabular}{cccc|cc}
\toprule
\textbf{Train}  & \textbf{Test} & \multicolumn{2}{c|}{\textbf{\mdeberta}} & \multicolumn{2}{c}{\textbf{\xlmroberta}} \\
                         &                        & \langid{EN}     & \langid{ES}     & \langid{EN}     & \langid{ES}     \\ \toprule
\multirow{2}{*}{No Met} & Met           & 73.86 \scriptsize{$\pm$ 3.10} & 73.38 \scriptsize{$\pm$ 1.92} & 77.21 \scriptsize{$\pm$ 3.05} & 75.61 \scriptsize{$\pm$ 1.30} \\
                        & No Met      & \textbf{75.65} \scriptsize{$\pm$ 2.10} & \textbf{73.89} \scriptsize{$\pm$ 1.11} & \textbf{78.50} \scriptsize{$\pm$ 2.80} & \textbf{77.13} \scriptsize{$\pm$ 1.92} \\   \midrule
\multirow{2}{*}{All} & Met           & 75.78 \scriptsize{$\pm$ 1.50} & 72.83 \scriptsize{$\pm$ 1.24} & 76.73 \scriptsize{$\pm$ 2.88} & 76.57 \scriptsize{$\pm$ 1.48} \\
                     & No Met       & \textbf{78.09}  \scriptsize{$\pm$ 0.56} & \textbf{75.92} \scriptsize{$\pm$ 1.73} & \textbf{79.31} \scriptsize{$\pm$ 2.39} & \textbf{79.23} \scriptsize{$\pm$ 1.73}           \\ 
\midrule
 average & & 75.84 & 74.01 & \textbf{77.94} & \textbf{77.13}\\
 \bottomrule
\end{tabular}
\end{table}

\section{Error Analysis}
In this section, we manually inspected a subset of erroneous cases in order to provide a qualitative insight into the results and with the intention of finding potential explanations of errors and models' performance for both detection and interpretation tasks.

\paragraph{\textbf{Detection}} We selected the predictions from the MLM that obtained highest F1 score from the monolingual experiments in \langid{ES} for both \meta and \cometa evaluations. We extracted false negatives and false positives and grouped the tokens by their number of occurrences (see Table \ref{tab:det_error_analysis_quanti}). 

\begin{table}[h!]
\caption{\textbf{Token-level error analysis for metaphor detection}: FP = False Positives (literal tokens predicted as metaphor), FN = False Negatives (missed metaphorical tokens). Results are grouped by test dataset and model.}
\label{tab:det_error_analysis_quanti}
\footnotesize
\begin{tabular}{@{}lllccc@{}}
\toprule
\textbf{Dataset} & \textbf{Model} & \textbf{Language} & \textbf{FP} & \textbf{FN} & \textbf{Total Errors} \\
\midrule
\multirow{1}{*}{\cometa} 
    & \xlmroberta & \langid{ES} & 132 & 175 & 307 \\
\midrule
\multirow{4}{*}{\meta} 
    & \mdeberta    & \langid{ES} & 137 & 262 & 399 \\
    & \xlmroberta & \langid{EN} & 409 & 360 & 769 \\
    & \llamasm       & \langid{ES} & 314 & 313 & 578 \\
    &  \gemma      & \langid{EN} & 444 & 551 & 995 \\
\bottomrule
\end{tabular}

\end{table}

Within the false positives of \cometa test set, we find tokens like \textit{apoyo} (lit. ``support'') or \textit{tensión} (lit. ``tension'') that appear more frequently used figuratively than with their literal sense or even only appear in the training set labeled as metaphor. Other wrong predictions are words that appeared in specific domains, such as texts that allude to the pandemic, e.g. \textit{ola} (lit. ``wave''), not detected as metaphorical due to its absence with metaphorical meaning in \meta sentences. 

Something similar occurs with the misclassified tokens from \meta test set. Most errors stem from conventional metaphors, namely \textit{gran}, \textit{abrir}, \textit{paso}, \textit{claro} (lit. ``great'', ``to open'', ``step'', ``clear'') that are regularly used with their metaphorical meaning. The lack of balanced examples might contribute to these predictions. However, we maintained the distribution as is, since our aim is to study the presence and prevalence of metaphor in natural language utterances.

We also conducted an error analysis of the LLMs experiments, focusing on the best-performing model \llamasm in \langid{ES} and \gemma in \langid{EN}. The observed error patterns show similar trends to those found in the MLMs, particularly in terms of false positive and false negative tokens, in Table \ref{tab:metaphor_error_analysis_tokens}. Given that \llamasm achieved a higher F1 score in \langid{ES}, we also examined cases where the model made correct predictions in \langid{ES} but failed to do so in \langid{EN}. Representative examples are provided in Table \ref{tab:det_quali_error_analysis}. As illustrated in examples 2 and 5, the \langid{EN} model makes a prediction, but it does not match the gold standard metaphor. In example 5, for instance, the model correctly identifies ``attack'' but misses ``hard''. In the remaining examples, the model fails to detect the metaphorical token entirely in \langid{EN}, despite successful identification in the \langid{ES} counterparts.

\begin{table}[ht]
\caption{\textbf{Error analysis of metaphor classification}: Most frequent false positives (FP) and false negatives (FN) by token, language, and frequency.}
\label{tab:metaphor_error_analysis_tokens}
\footnotesize
\begin{tabular}{lll}
\toprule
\textbf{Language} & \textbf{Type} & \textbf{Tokens (Frequency)} \\
\midrule
\multirow{2}{*}{\langid{ES}} 
 & FN & \textit{gran} (6), \textit{limpio} (3), \textit{claro} (3), \textit{gran} (2), \textit{fusionarán} (2), \textit{fuerzas} (2) \\
 & FP & \textit{gran} (4), \textit{postura} (2), \textit{paso} (2), \textit{llegaron} (2), \textit{conexiones} (2), \textit{claro} (2) \\
\multirow{2}{*}{\langid{EN}} 
 & FN & \textit{great} (6), \textit{forces} (4), \textit{placed} (3), \textit{surrounding} (2), \textit{see} (2), \textit{scattered} (2) \\
 & FP & \textit{great} (5), \textit{lead} (4), \textit{high} (4), \textit{ways} (3), \textit{support} (3), \textit{reach} (3) \\
\bottomrule
\end{tabular}

\end{table}

\begin{table}[ht]
\caption{\textbf{Examples correctly predicted in Spanish but missed in English}. Metaphorical tokens in Spanish are in bold. In English, the labeled metaphors are in column \textit{Gold (EN)} and predicted metaphors in \textit{Pred. (EN)}.}
\label{tab:det_quali_error_analysis}
\footnotesize
\renewcommand{\arraystretch}{1.3}
\resizebox{\linewidth}{!}{
\begin{tabular}{>{\bfseries}r p{4.5cm} p{4.5cm} p{1.8cm} p{1.8cm}}
\toprule
\# & \textbf{English Sentence} & \textbf{Spanish Sentence} & \textbf{Gold (\langid{EN})} & \textbf{Pred. (\langid{EN})} \\
\midrule
1 & The social relationship of forces is inevitable. & La relación social de \textbf{fuerzas} es algo inevitable. & forces & \textemdash{} \\
2 & The right-wing parties have provided a link to positively orient the vote of the population. & Los partidos de derechas han facilitado un link para \textbf{orientar} positivamente el voto de la población. & orient & link \\
3 & If there is a surge of electricity, it's very dangerous to anyone around. & Si hay una \textbf{oleada} de electricidad, es muy peligroso para cualquiera que esté cerca. & surge & \textemdash{} \\
4 & It's hard to install the system because hackers attack it every night. & Es difícil instalar el sistema porque los piratas informáticos lo \textbf{atacan} todas las noches. & attack & hard \\
\bottomrule
\end{tabular}
}

\end{table}

\paragraph{\textbf{Interpretation}}

We analysed a subset of 30 errors from each experimental setup, both \langid{EN} and \langid{ES}, and evaluation sets. We chose the predictions from \xlmroberta since it is the model that performs better in this task in most scenarios. Although results show that models struggle more to identify the inference relation if there is a metaphorical expression involved in the pair sentences, we do not observe any particular feature within the errors. 

A remarkable aspect to highlight is the lower results of \esxnli with respect to \xnlidev and \xnlitest in Table \ref{tab:xnli_interpretation_evaluation}. It could be motivated by the difference in the text domains since \xnli is an extension of MultiNLI, which maintains the same set of textual domains. While \esxnli is a collection of texts from another set of genres and sources. 
Regarding \langid{EN}, some of the errors might derive from the misclassification of some pairs, since annotations were developed on \langid{ES} text. A metaphorical expression involved in the inference relationship in \langid{ES} might not be present in its \langid{EN} version and vice versa. Thus, samples from these two classes, pairs with and without metaphors, should be reexamined in \langid{EN} and correctly classified for future experimentation. 
 
\section{Conclusions and Future Work}

In this work, we present \meta, the first cross-lingual parallel dataset in \langid{ES} and \langid{EN} labeled for metaphor detection and interpretation framed within the task of NLI. This new resource allows to perform a series of experiments to assess the capabilities of MLMs and LLMs when dealing with this kind of figurative expressions present in natural language utterances.

Regarding the task of metaphor detection, after the annotation process and experimental results, we can conclude the importance of establishing a unified criterion for annotation that is valid for different languages if the aim is to continue researching cross-lingual approaches. In addition, the semi-automatic process of annotation followed for \langid{EN} shows that automatic labeling of cross-lingual metaphor is far from trivial. Metaphorical expressions are language and culture-dependent. Moreover, the translation of the data introduces a new layer in which metaphorical expressions can either be lost from the translation of the source to the target language, or can be introduced in the target language by means of the translator, either human or automatic. Further work to explore automatic annotation methodologies would be of considerable value in reducing the demanding workload and effort of manual labeling in more than one language. 

With respect to results, the purpose of our experiments is not to outperform the state-of-the-art results, but to analyze LMs' capabilities in processing metaphorical language in a multilingual and crosslingual setting. Best results are obtained when \meta in both languages is used for training. The augmentation of the training set size and the parallel annotations might boost this performance. Cross-domain and monolingual experiments show how the lack of consistency in the annotation criteria affects the performance of models. This can be observed as well in the zero-shot cross-lingual setup, although the scenario of training in \langid{EN} and evaluating in \langid{ES} shows competitive performance. It should be noted that the \langid{EN} set contains a larger number of instances annotated as metaphorical in the training set. In addition, the in-vocabulary and out-of-vocabulary evaluation points to some kind of bias in the learning process. This could stem from the fact that the majority of the metaphor instances are conventional or due to \textit{lexical memorization} \cite{ levy-etal-2015-supervised, boisson2023construction}. Future research on this line of work should be carried out to clarify this issue.

Regarding metaphor interpretation, we evaluated the ability of MLMs and LLMs to understand metaphorical expressions framed within the NLI task. We provide parallel annotations at the premise-hypothesis pair level that mark whether the presence of metaphorical expressions is relevant for the inference relationship.  We exploited this information to conduct our experiments. From the reported results, we can observe a tendency of the models to perform lower with pairs that contain at least one metaphorical expression. However, this trend breaks when evaluating the datasets in their translated version. We presume the translation process might induce biases in metaphor occurrence and the ``naturalness'' of the sentences. Similarly to metaphor detection, future work to analyse the impact of translation in the development of metaphor parallel resources should be explored for the task of metaphor interpretation, as well as additional experimentation from a multilingual perspective.

In summary, our work provides a high-quality, cross-lingual and parallel resource with aligned annotations for detection and interpretation over the same text. Our new dataset not only facilitates systematic evaluation of model performance across languages but also serves as a starting point for future research in metaphor detection, interpretation, and metaphor transfer across languages. By tackling both annotation and empirical challenges, we lay the groundwork for more accurate and critical assessments of how language models handle metaphor and meaning across languages.

\section{Limitations}
Metaphor annotation is an inherently subjective task. This variance in annotations is reflected in \metaen, due to the different criteria employed through the annotation process. Labels in this language should be updated and further revised to improve their quality. Disagreement and subjectivity could be counterbalanced by more annotation iterations and a larger number of annotators, in order to develop more consistent and reliable labeled data. However, this constitutes an arduous and costly process. Data augmentation and semi-automatic methods could be exploited to create larger datasets with similar characteristics to the one we present and extend it to more languages, since most corpora available for metaphor processing are of reduced size and limited to a narrow set of languages. The existence of parallel resources in multiple languages other than \langid{EN} that reflect cultural and real-world knowledge nuances is of great importance to continue researching such a complex phenomenon as figurative language, specifically metaphors. 

\clearpage

%\section{Appendix} \label{sec:appendix}

%\onecolumn

\appendix

\appendixsection{Data Splits Statistics}
%\label{sec:appendixA}

% Please add the following required packages to your document preamble:
% \usepackage{multirow}
\begin{table}[h!]
\caption{Number of tokens annotated as metaphors and sentences that contain at least one metaphorical expression in each data split for metaphor detection experiments in \langid{ES}.}
\label{tab:es_det_datasplits_analysis}
\footnotesize
\begin{tabular}{crcc|rcc}
\toprule
                                                 & \multicolumn{3}{c|}{Tokens}     & \multicolumn{3}{c}{Sentences}  \\ 
                                                 & Metaphor & Total  & Metaphor \% & Metaphor & Total & Metaphor \% \\ \midrule
{Train}  & 1053  & 79462 & 1.33 & 893 & 7259 & 12.30\ \\
{Dev} & 474 & 32285 & 1.47 & 364 & 2431 & 14.97 \\
{Test}   & 767  & 52892 & 1.63 & 616 & 3630 & 16.97\ \\ \midrule
{Total}    &                    2294     & 164639 & 1.39      & 1873     & 13320 & 14.06 \\
\bottomrule
\end{tabular}

\end{table} 
% Please add the following required packages to your document preamble:
% \usepackage{multirow}
\begin{table}[h!]
\caption{Number of tokens annotated as metaphors and sentences that contain at least one metaphorical expression in each data split for metaphor detection experiments in \langid{EN}.}
\label{tab:en_det_datasplits_analysis}
\footnotesize
\begin{tabular}{crcc|rcc}
\toprule
                                                 & \multicolumn{3}{c|}{Tokens}     & \multicolumn{3}{c}{Sentences}  \\ 
                                                 & Metaphor & Total  & Metaphor \% & Metaphor & Total & Metaphor \% \\ \midrule
{Train}  & 1527  & 75935 & 2.01 & 1285 & 7287 & 17.63\ \\
{Dev} & 697 & 30733 & 2.27 & 553 & 2422 & 22.83 \\
{Test}   & 1106  & 50153 & 2.21 & 898 & 3611 & 24.87\ \\ \midrule
{Total}    &                    3330     & 156821 & 2.12      & 2736     & 13320 & 20.54 \\
\bottomrule
\end{tabular}
\end{table} 
% Please add the following required packages to your document preamble:
% \usepackage{multirow}
\begin{table}[h!]
\caption{Number of premise-hypothesis pairs with and without metaphorical expressions in data splits used for metaphor interpretation experiments, with results in \ref{tab:xnli_interpretation_finetune}. Non-relevant (\textit{non-rel}) cases were not exploited due to their ambiguity.}
\label{tab:interpretation_datasplits_analysis}
\footnotesize
\begin{tabular}{crcrcrcr}
\toprule                                 & \multicolumn{7}{c}{Premise-Hypothesis Pairs}                                               \\
                            &    Met & Met \% & Non-rel & Non-rel\% & No met & No met\% & Total \\ \midrule
{Train}  &  796      & 12.46     & 1109    & 17.37       & 4481        & 70.17        & 6386    \\ \midrule
{Dev} &  201     & 12.55     & 278        & 17.35       & 1123        & 70.10        & 1602  \\ \midrule
{Test}     &  251      & 12.54     & 348        & 17.38       & 1403        & 70.08        & 2002    \\ \midrule
Total                      & 1248     & 12.49     & 1735       & 17.37       & 7007        & 70.14        & 9990 \\ \bottomrule
\end{tabular}
\end{table} 

\clearpage

\appendixsection{Inter-Annotator Agreement Labeling Process}

\begin{figure}[h!]
    \includegraphics[height=0.85\textheight]{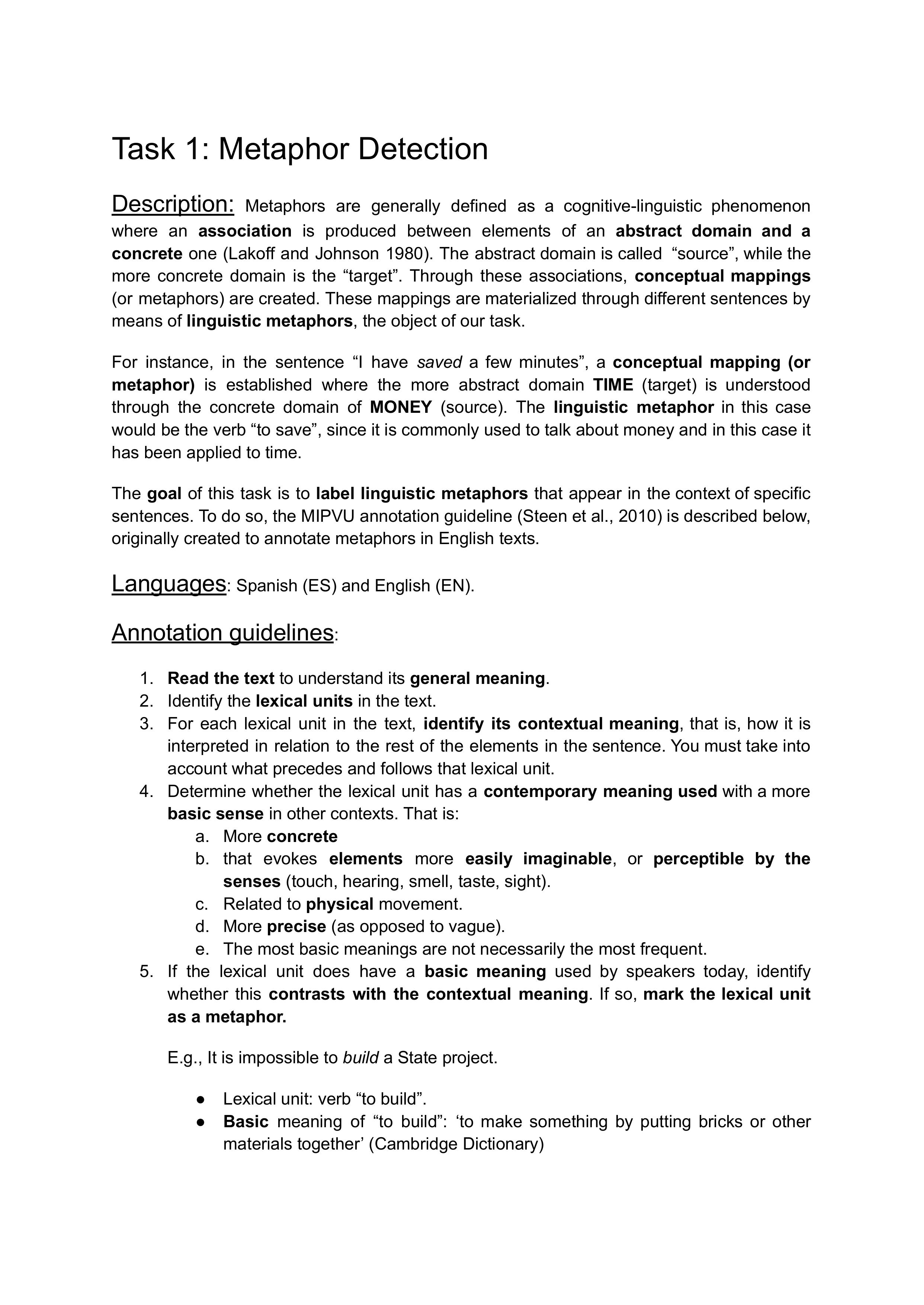}
\end{figure}

\begin{figure}[h!]
    \includegraphics[height=0.90\textheight]{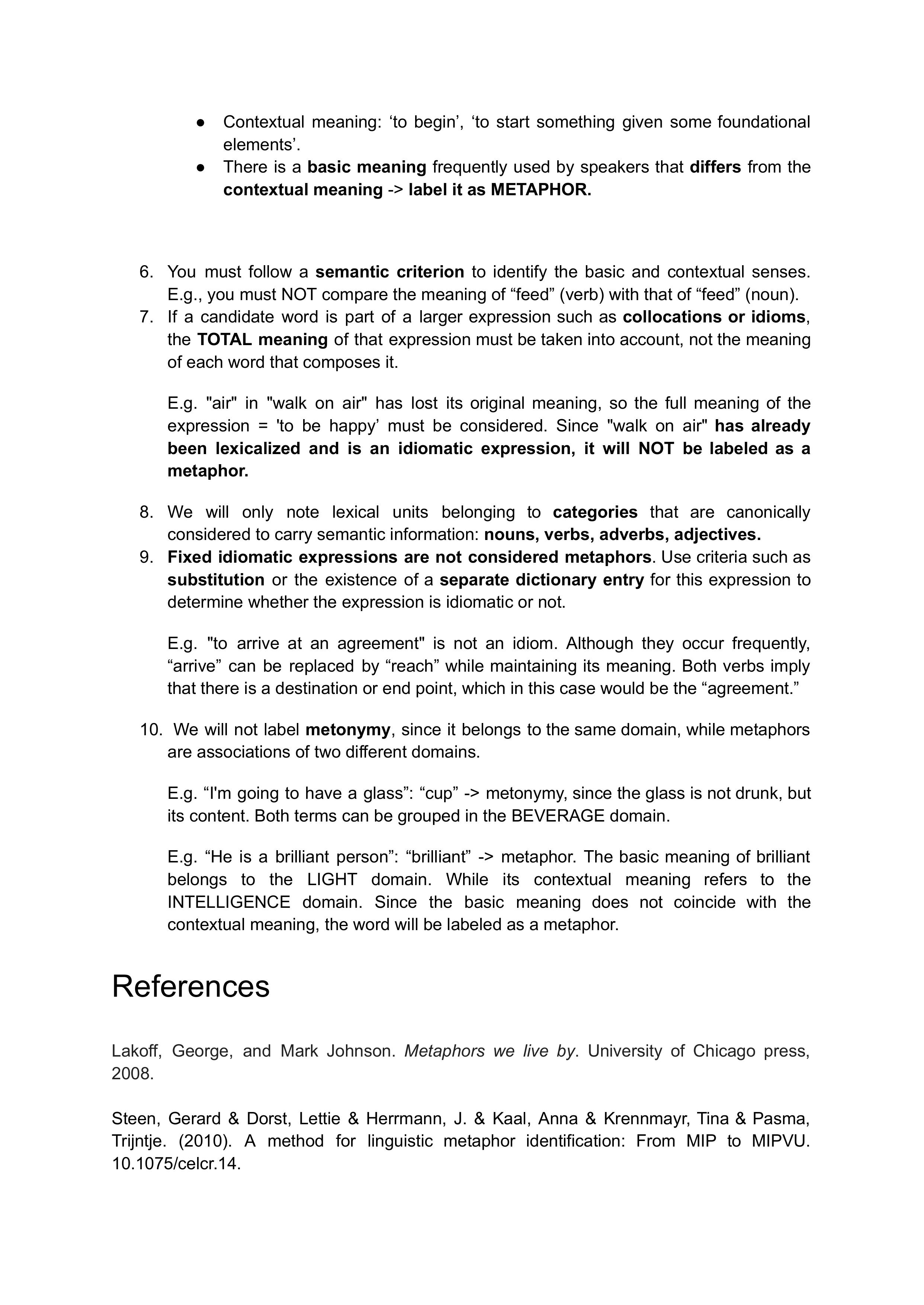}
    \caption{Guidelines for metaphor detection provided to the two annotators.}
    \label{fig:annotation-guidelines}
\end{figure}

\begin{figure}[h!]
    \includegraphics[width=0.9\linewidth]{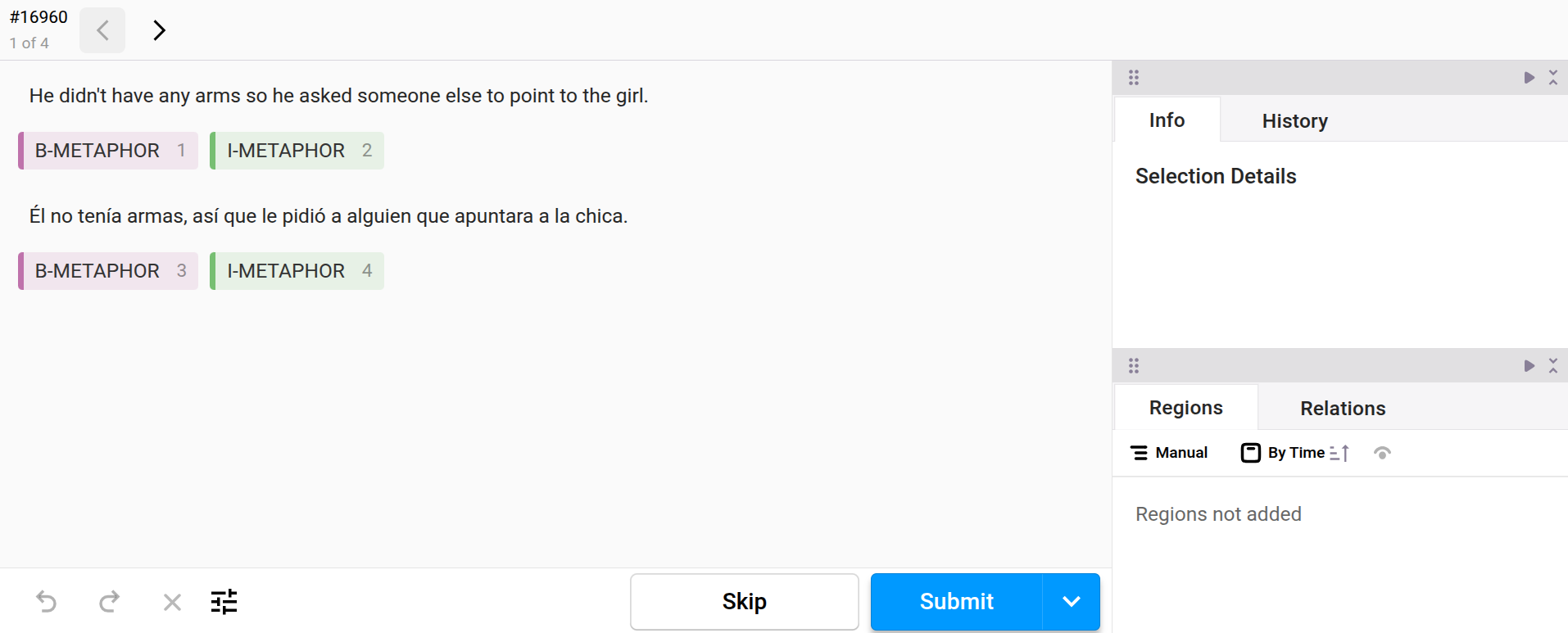}
    \caption{Image of the platform Label Studio \cite{Label_Studio} used by the annotators.}
    \label{fig:enter-label}
\end{figure}
\clearpage

\appendixsection{Detection Experiments Results} 

% Please add the following required packages to your document preamble:
% \usepackage{multirow}
\begin{table}[h!]
\caption{\textbf{Cross-domain metaphor detection: } F1, precision and recall scores from evaluation of \meta on models trained with other metaphor detection datasets of different textual domain. Best model performance in bold.}
\footnotesize
\label{tab:appendix_xnli_det_crossdomain}
\begin{tabular}{c|c|ccc|ccc}
\toprule
\multirow{3}{*}{\textbf{Train (\langid{ES})}} & \multirow{3}{*}{\textbf{Test (\langid{ES})}} & \multicolumn{3}{c|}{\multirow{2}{*}{\textbf{\mdeberta}}}      & \multicolumn{3}{c}{\multirow{2}{*}{\textbf{\xlmroberta}}} \\
&   & \multicolumn{3}{c|}{} & \multicolumn{3}{c}{} \\
                        &                  & F1 & Prec.   & Rec.  & F1   & Prec.   & Rec.  \\ \midrule
\multirow{11}{*}{\cometa} & \esxnli all      & \textbf{64.94} & 77.12 & 56.08 & 60.52 & 75.52  & 50.49    \\
                        & \esxnli hyp      & \textbf{58.78} & 69.43 & 50.96 & 55.58 & 71.36 & 45.51    \\
                        & \esxnli prem     & \textbf{69.74} & 83.16 & 60.05 & 64.22 & 78.49  & 54.34  \\ \cmidrule{2-8} 
                        & \xnlidev all   & \textbf{62.61} & 76.50 & 52.98 & 58.11 & 73.82  & 47.92    \\
                        & \xnlidev hyp   & \textbf{63.49} & 74.63 & 55.23 & 59.49 & 71.57  & 50.90    \\
                        & \xnlidev prem  & \textbf{61.67} & 78.65 & 50.72 & 56.62 & 76.54  & 44.93    \\ \cmidrule{2-8} 
                        & \xnlitest all  & \textbf{60.11} & 71.24 & 52.00 & 58.46 & 73.14 & 48.69    \\
                        & \xnlitest hyp  & \textbf{57.89} & 68.56 & 50.10 & 55.63 & 71.39 & 45.57    \\
                        & \xnlitest prem & \textbf{62.70} & 74.35 & 54.20 & 61.63 & 75.00 & 52.31    \\ \cmidrule{2-8}
                        & average & 62.44 & 74.85 & 53.59 & 58.92 & 74.09 & 48.96 \\ \cmidrule{2-8} 
                        & \cometa & 67.46 & 78.70 & 59.03 & 67.44 & 75.57 & 60.88 \\ 
                        \bottomrule
\end{tabular}

\begin{tabular}{c|c|ccc|ccc}
\multirow{3}{*}{\textbf{Train (\langid{EN})}} & \multirow{3}{*}{\textbf{Test (\langid{EN})}} & \multicolumn{3}{c|}{\multirow{2}{*}{\textbf{DeBERTa}}} & \multicolumn{3}{c}{\multirow{2}{*}{\textbf{\xlmroberta}}} \\
                        &                  & \multicolumn{3}{c|}{} & \multicolumn{3}{c}{} \\
                        &                  & F1   & Prec.  & Rec.  & F1   & Prec.   & Rec.  \\ \midrule
\multirow{11}{*}{\vuam}   & \esxnli all      & \textbf{38.02} & 24.96 & 79.75 & 35.23  & 22.97  & 75.53    \\
                        & \esxnli hyp      & \textbf{32.63} & 20.48 & 80.30 & 30.60  & 19.11  & 76.85    \\
                        & \esxnli prem     & \textbf{43.46} & 29.92 & 79.34 & 39.88  & 27.22  & 74.54    \\ \cmidrule{2-8} 
                        & \xnlidev all   & \textbf{37.52} & 24.75 & 77.46 & 34.67  & 22.35  & 77.22    \\
                        & \xnlidev hyp   & \textbf{35.69} & 22.87 & 81.16 & 33.02  & 20.89  & 78.84    \\
                        & \xnlidev prem  & \textbf{40.09} & 27.59 & 73.30 & 36.83  & 24.37  & 75.39    \\ \cmidrule{2-8} 
                        & \xnlitest all  & \textbf{34.37} & 22.18 & 76.31 & 32.09  & 20.28  & 76.74     \\
                        & \xnlitest hyp  & \textbf{31.70} & 20.12 & 74.70 & 30.18  & 18.81  & 76.32    \\ 
                        & \xnlitest prem & \textbf{38.13} & 25.20 & 78.87 & 34.75  & 22.42  & 77.25 \\ \cmidrule{2-8}
                        & average & 36.85 & 24.23 & 77.91 & 34.14 & 22.05 & 76.52 \\  \cmidrule{2-8}
                        & \vuam & 73.79 & 79.95 & 68.50 & 72.11  & 73.95  & 70.36 \\ 

    \bottomrule
\end{tabular}
\end{table}
\begin{table}[h!]
\caption{
\textbf{Monolingual metaphor detection with encoder-only  models}: F1, precision and recall results from model fine-tuning with \meta (M4X) and evaluation with its test set and \vuam (\langid{EN}) and \cometa (\langid{ES}) test sets, for each corresponding language. Scores are an average of results from 5 random runs, standard deviation next to F1 scores.
}
\label{tab:appendix_xnli_det_finetune}
\footnotesize
\begin{tabular}{cc|ccc|ccc}
\toprule
    \multicolumn{2}{c}{} & \multicolumn{3}{c}{\textbf{\mdeberta}} & \multicolumn{3}{c}{\textbf{\xlmroberta}} \\
    \textbf{Train} & \textbf{Test} & F1 & Prec. & Rec. & F1 & Prec. & Rec.  \\
     \midrule
     \multirow{2}{*}{M4X\scriptsize{\langid{ES}}}  & 
            M4X\scriptsize{\langid{ES}} &  \textbf{67.17} \scriptsize{$\pm$ 0.72} & 73.60 \scriptsize{$\pm$ 5.38} & 62.24 \scriptsize{$\pm$ 4.38} & 66.84 \scriptsize{$\pm$ 0.87} & 73.29 \scriptsize{$\pm$ 2.35} &  61.59 \scriptsize{$\pm$ 3.17} \\
            & \cometa & 55.62 \scriptsize{$\pm$ 0.61} & 61.14 \scriptsize{$\pm$ 5.05} & 51.53 \scriptsize{$\pm$ 4.32} & \textbf{56.24} \scriptsize{$\pm$ 1.28} & 59.32 \scriptsize{$\pm$ 1.75} & 53.57 \scriptsize{$\pm$ 2.74} \\
            \midrule
     \multirow{2}{*}{M4X\scriptsize{\langid{EN}}}  & 
            M4X\scriptsize{\langid{EN}} &  58.02 \scriptsize{$\pm$ 0.95} & 65.82 \scriptsize{$\pm$ 2.83} & 52.09 \scriptsize{$\pm$ 3.00} & \textbf{59.24}  \scriptsize{$\pm$ 1.00} &  63.70 \scriptsize{$\pm$ 3.40} & 56.31 \scriptsize{$\pm$ 4.19} \\
            & \vuam & 29.14 \scriptsize{$\pm$ 1.63} & 76.69 \scriptsize{$\pm$ 1.84} & 18.02 \scriptsize{$\pm$ 1.37} & \textbf{32.42} \scriptsize{$\pm$ 2.68} & 76.70 \scriptsize{$\pm$ 3.08} & 20.68 \scriptsize{$\pm$ 2.48}  \\
    \bottomrule
\end{tabular}
\end{table}

\begin{table*}[h!]
\caption{
\textbf{Monolingual metaphor detection with decoder-only models}: F1, precision and recall results from model fine-tuning with \meta (M4X) and evaluation with its test set and \vuam (\langid{EN}) and \cometa (\langid{ES}) test sets, for each corresponding language. Best model for each test set in bold.
} \label{tab:appendix_xnli_det_finetune_llms}
\footnotesize
\begin{tabular}{cc|ccc|ccc|ccc} 
\toprule
    \multicolumn{2}{c}{} & \multicolumn{3}{c}{\textbf{\llamasm}} & \multicolumn{3}{c}{\textbf{\qwensm}} &  \multicolumn{3}{c}{\textbf{\gemma}}\\
    \textbf{Train} & \textbf{Test} & F1 & Prec. & Rec. & F1 & Prec. & Rec.   & F1 & Prec. &Rec.   \\
     \midrule
     \multirow{2}{*}{M4X\scriptsize{\langid{ES}}}  & 
            M4X\scriptsize{\langid{ES}} &  \textbf{59.15}& 59.11& 59.19& 56.76& 65.98&   49.80& 58.06& 65.47&52.15\\
            & \cometa & \textbf{50.06}& 53.56& 46.99& 44.26& 53.07&  37.96& 45.32& 52.60&39.81\\
            \midrule
     \multirow{2}{*}{M4X\scriptsize{\langid{EN}}}  & 
            M4X\scriptsize{\langid{EN}} &  50.41& 53.68& 47.52& 47.66&  44.79&  50.92& \textbf{52.61}& 55.50&50.00\\
            & \vuam & \textbf{26.78}& 61.68& 17.10& 26.44& 59.72&  16.98& 26.72& 70.19&16.50\\
    \bottomrule
\end{tabular}
\end{table*}

% Required packages:
% \usepackage{booktabs, multirow}
\begin{table}[h!]
\caption{\textbf{Multilingual metaphor detection:} F1, precision, and recall for each model trained on \metaes+\metaen, evaluated on Spanish and English test sets. Best model (encoder-only and decoder-only) for each test set in bold.}
\label{tab:aoo_multilang_det}
\footnotesize
\begin{tabular}{lccc|ccc}
\toprule
 & \multicolumn{3}{c}{\textbf{\metaes}} & \multicolumn{3}{c}{\textbf{\metaen}} \\
 & F1 & Prec. & Rec. & F1 & Prec. & Rec. \\
\midrule
\textbf{\mdeberta}     & \textbf{68.70} \tiny{$\pm$ 1.27} & 72.78 \tiny{$\pm$ 3.45} & 65.29 \tiny{$\pm$ 3.77} & \textbf{60.80} \tiny{$\pm$ 1.05} & 65.03 \tiny{$\pm$ 2.08} & 57.23 \tiny{$\pm$ 3.10} \\
\textbf{\xlmroberta}   & 66.01 \tiny{$\pm$ 0.55} & 70.35 \tiny{$\pm$ 4.42} & 62.53 \tiny{$\pm$ 3.93} & 60.14 \tiny{$\pm$ 1.32} & 64.97 \tiny{$\pm$ 2.81} & 56.15 \tiny{$\pm$ 3.04} \\ \midrule
\textbf{\llamasm}      & \textbf{61.46} & 61.90 & 61.02 & \textbf{55.26} & 53.33 & 57.34 \\
\textbf{\qwensm}       & 58.72 & 59.31 & 58.15 & 52.62 & 47.56 & 58.90 \\
\textbf{\gemma}        & 57.02 & 64.89 & 50.85 & 54.60 & 57.79 & 51.74 \\
\bottomrule
\end{tabular}
\end{table}

\begin{table}[h!]
\caption{
\textbf{Zero-shot cross-lingual metaphor detection}: F1, precision and recall scores of models performance after fine-tuning with \metaes and testing on \metaen, and vice versa. Scores are an average of 5 random runs, standard deviation next to F1 scores. Best model performance for each evaluation in bold.
}

\label{tab:appendix_xnli_det_crosslingual}
\footnotesize
\begin{tabular}{cc|ccc|ccc} 
\toprule
    \multicolumn{2}{c}{} & \multicolumn{3}{c}{\textbf{mDeBERTa}} & \multicolumn{3}{c}{\textbf{XLM-RoBERTa}} \\
    \textbf{Train} & \textbf{Test} & F1 & Prec. & Rec. & F1 & Prec. & Rec.  \\
     \midrule
     \langid{ES}  &  \langid{EN} &  37.78 \scriptsize{$\pm$ 4.35} & 71.10 \scriptsize{$\pm$ 4.43} & 26.09 \scriptsize{$\pm$ 5.15} & \textbf{39.90} \scriptsize{$\pm$ 2.20} & 69.77 \scriptsize{$\pm$ 2.22} &  28.02 \scriptsize{$\pm$ 2.50} \\
     \midrule
     \langid{EN}  & \langid{ES} &  56.41 \scriptsize{$\pm$ 3.05} & 62.69 \scriptsize{$\pm$ 3.51} & 51.63 \scriptsize{$\pm$ 5.39} & \textbf{58.94}  \scriptsize{$\pm$ 1.32} &  60.20 \scriptsize{$\pm$ 6.14} & 58.77 \scriptsize{$\pm$ 6.66} \\
            
    \bottomrule
\end{tabular}
\end{table}

\clearpage

\appendixsection{Prompts for Inference on Interpretation}
%\label{sec:appendix_prompts}

\begin{table}[h!]
\small
\caption{
Prompts used in the evaluation of LLMs for metaphor interpretation via NLI.}\label{tab:eval-prompts}
\begin{tabular}{p{0.45\textwidth}}
\toprule
\textbf{Zero-shot} \\
\midrule
Say which is the inference relationship between these two sentences. Please, answer only with one word between 'entailment', 'neutral 'or 'contradiction'. \\
\{Premise\} -> \{Hypothesis\}: \\
\midrule
\textbf{Chain-of-thought} \\
\midrule
You are an expert linguist and your task is to annotate sentences for the task of Natural Language Inference. This task consists in determining if a first sentence (premise) entails, contradicts or does not entail nor contradict the second sentence (hypothesis). Please, answer with one of the following labels: 'entailment', 'contradiction' or 'neutral'. \\ 

Here you have a few examples: \\ \\
Premise: I am an open person. \\
Hypothesis: I am friendly. \\
Answer: entailment \\ \\

Premise: My heart is broken. \\
Hypothesis: I am happy. \\
Answer: contradiction. \\ \\

Premise: You are a close friend. \\
Hypothesis: I like your eyes. \\
Answer: neutral.\\ \\

\{Premise\}:\\ 
\{Hypothesis\}: \\
\{Answer\}:\\ 
\bottomrule
\end{tabular}
\end{table}

\starttwocolumn

\begin{acknowledgments}
This work has been supported by the HiTZ center and the Basque Government
(Research group funding IT-1805-22). Elisa Sanchez-Bayona is funded by the UPV/EHU PIF20/139 grant. We also thank the funding from
the following MCIN/AEI/10.13039/501100011033 projects: (i) DeepKnowledge
(PID2021-127777OB-C21) and ERDF A way of making Europe; (ii) Disargue
(TED2021-130810B-C21) and European Union NextGenerationEU/PRTR; (iii)  DeepMinor (CNS2023-144375) and European Union NextGenerationEU/PRTR.
\end{acknowledgments}

\bibliographystyle{compling}
\bibliography{COLI_template, conll_papers}

\end{document}